\documentclass[pdflatex,sn-mathphys-num]{sn-jnl} 

\usepackage{graphicx}
\usepackage{multirow}
\usepackage{amsmath,amssymb,amsfonts}
\usepackage{amsthm}
\usepackage[title]{appendix}
\usepackage{xcolor}
\usepackage{textcomp}
\usepackage{manyfoot}
\usepackage{booktabs}
\usepackage{algorithm}
\usepackage{algorithmicx}
\usepackage{algpseudocode}
\usepackage{listings}
\usepackage{subcaption}

\begin{document}

\title[Beyond the Mean]{Beyond the Mean: Distribution-Aware Loss Functions for Bimodal Regression}


\author*[1]{\fnm{Abolfazl} \sur{Mohammadi-Seif}}\email{abolfazl.mohammadiseif@upf.edu}
\author[2]{\fnm{Carlos} \sur{Soares}}\email{csoares@fe.up.pt}
\author[3,4]{\fnm{Rita P.} \sur{Ribeiro}}\email{rpribeiro@fc.up.pt}
\author[1]{\fnm{Ricardo} \sur{Baeza-Yates}}\email{ricardo.baeza@upf.edu}

\affil*[1]{\orgdiv{Department of Engineering}, \orgname{Universitat Pompeu Fabra}, \orgaddress{\street{Roc Boronat St.}, \city{Barcelona}, \postcode{08018}, \state{Barcelona}, \country{Spain}}}

\affil[2]{\orgdiv{Department of Engineering}, \orgname{University of Porto}, \orgaddress{\street{Praça de Gomes Teixeira}, \city{Porto}, \postcode{4099-002}, \country{Portugal}}}

\affil[3]{\orgdiv{Faculty of Sciences}, \orgname{University of Porto}, 
\orgaddress{\postcode{4169-007} \city{Porto}, \country{Portugal}}}

\affil[4]{\orgname{INESC TEC}, 
\orgaddress{\postcode{4200-465} \city{Porto}, \country{Portugal}}}


\abstract{
Despite the strong predictive performance achieved by machine learning models across many application domains, assessing their trustworthiness through reliable estimates of predictive confidence remains a critical challenge. This issue arises in scenarios where the likelihood of error inferred from learned representations follows a bimodal distribution, resulting from the coexistence of confident and ambiguous predictions. Standard regression approaches often struggle to adequately express this predictive uncertainty, as they implicitly assume unimodal Gaussian noise, leading to mean-collapse behavior in such settings. Although Mixture Density Networks (MDNs) can represent different distributions, they suffer from severe optimization instability.
We propose a family of distribution-aware loss functions integrating normalized RMSE with Wasserstein and Cramér distances. When applied to standard deep regression models, our approach recovers bimodal distributions without the volatility of mixture models. Validated across four experimental stages, our results show that the proposed Wasserstein loss establishes a new Pareto efficiency frontier: matching the stability of standard regression losses like MSE in unimodal tasks while reducing Jensen-Shannon Divergence by 45\% on complex bimodal datasets. Our framework strictly dominates MDNs in both fidelity and robustness, offering a reliable tool for aleatoric uncertainty estimation in trustworthy AI systems.
}

\keywords{Bimodal Regression, Distributional Loss, Trustworthy AI}

\maketitle


\section{Introduction}\label{sec:intro}

Deep learning models, such as EfficientNet, have achieved remarkable success in image classification. However, a model's confidence is often not calibrated to its actual correctness. To build trustworthy systems, we often need a secondary \textit{complexity predictor} that estimates the likelihood of error for a given input based on its embeddings. Standard approaches typically treat this complexity as a continuous scalar value to be regressed, implicitly assuming a unimodal Gaussian distribution of difficulty.

In practice, this assumption is not well supported. As we show in our analysis, the distribution of prediction errors is not Gaussian but exhibits a clear bimodal shape, reflecting a split between confidently correct and systematically hard samples. This mismatch suggests that regression-based complexity predictors may be inherently mis-specified, motivating the need for a more faithful modeling approach.

To better understand whether this assumption holds in practice, we conducted a preliminary investigation using standard image classification benchmarks. We trained an EfficientNet-V2-M model \cite{tan2019efficientnet} on the CIFAR-100 and Food-101 datasets using standard cross-entropy loss. Post-training, we computed the \textit{prediction error} for each sample $x_i$ with true class label $c_i$. We denote this prediction error by $y_i$, defined as the complement of the model's predicted probability assigned to the ground truth class:

\begin{equation}
    y_i = 1 - \hat{p}(c_i \mid x_i)
\end{equation}

\begin{figure}[ht]
    \centering
    \begin{subfigure}[b]{0.48\textwidth}
        \includegraphics[width=\textwidth]{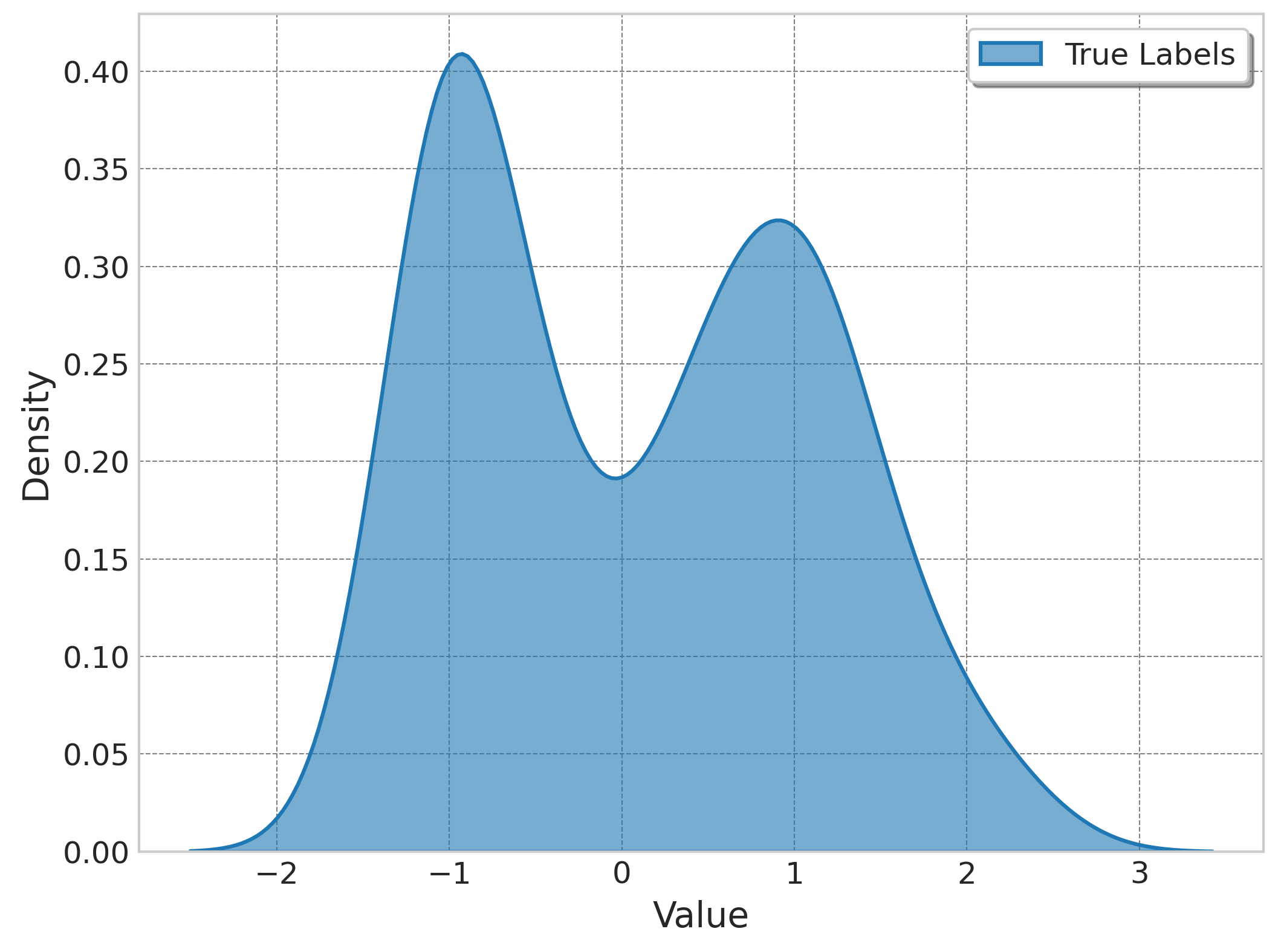}
        \caption{}
    \end{subfigure}
    \hfill
    \begin{subfigure}[b]{0.48\textwidth}
        \includegraphics[width=\textwidth]{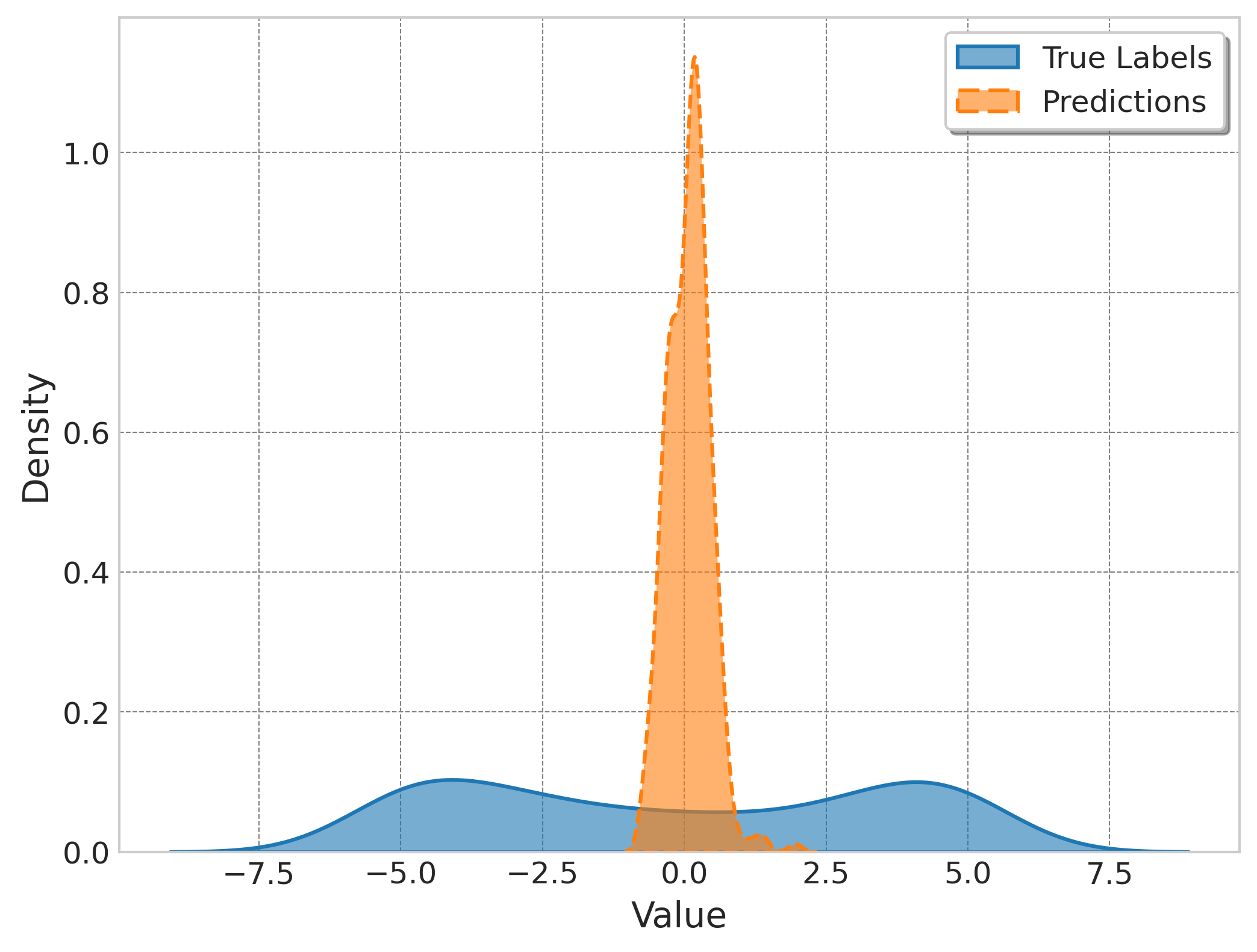}
        \caption{}
    \end{subfigure}
    \vspace{0.2cm}
    \caption{Bimodal distribution of targets (a). Standard regression models trained with MSE fail to capture the underlying bimodal distribution (b).}
    \label{fig:Fig0_bimodal_dist_pred_collaps}
\end{figure}

Analysis of the distribution of these $y_i$ values reveals a distinct bimodal shape rather than a Gaussian distribution (Fig.~\ref{fig:Fig0_bimodal_dist_pred_collaps}a). The modes correspond to two regimes:
\textit{easy}, where the classifier is correct and confident  
($\hat{p} \approx 1, e \approx 0$), and \textit{hard}, where the classifier misclassifies or has low confidence ($
\hat{p} \ll 1, e > 0$).

This empirical observation forms the primary motivation for our work. As illustrated in Figure~\ref{fig:Fig0_bimodal_dist_pred_collaps} (b), standard regression models trained with Mean Squared Error (MSE) fail to capture this bimodal reality. Because MSE minimizes the squared difference from the mean, the model's predictions collapse to a single mode in the center, the sparse region between the two actual clusters. Consequently, the model predicts a state of \textit{average difficulty} that rarely exists in practice, rather than identifying the distinct easy or hard nature of the samples.

To bridge this gap, we propose a distribution-aware loss framework. By integrating normalized RMSE with statistical distance metrics, we treat the regression target not as a single point, but as a continuous probability measure. This formulation compels the model to align its predictive distribution with the true target distribution, theoretically enabling the recovery of bimodal distribution without requiring complex architectural changes. We validate this framework through a comprehensive four-stage experimental protocol to rigorously test its stability and fidelity.

We formulate the following research questions to guide our investigation.

\begin{description}
    \item[\bfseries RQ1 (Fidelity):] Can distance-based loss functions recover bimodal error distributions where standard point-estimate losses suffer from mode collapse?
    \item [\textbf{RQ2 (Stability):}] How do these losses behave during the transition from simple to complex noise distributions? Can they avoid the optimization instability often observed in mixture models?
    \item[\textbf{RQ3 (Robustness):}] Are the proposed losses robust when applied to real-world tabular tasks with unknown target distributions?
    \item[\textbf{RQ4 (Application):}] In the motivating context of instance difficulty assessment, can the framework accurately distinguish between \textit{easy} and \textit{hard} samples?
\end{description}






The remainder of this paper is organized as follows. Section \ref{sec:related} reviews related work. Section \ref{sec:method} details the proposed loss framework and its mathematical formulation. Section \ref{sec:setup} presents the four-stage experimental evaluation, ranging from synthetic benchmarks to high-dimensional computer vision tasks, while Section \ref{sec:results} presents the results. Section \ref{sec:discussion} discusses the implications of the findings regarding the stability-fidelity trade-off. Finally, Section \ref{sec:conclusion} concludes the study and outlines future directions.

\section{Related Work}\label{sec:related}

This research lies at the intersection of three domains: probabilistic modeling of aleatoric uncertainty, regression techniques for imbalanced or non-parametric distributions, and the motivating application of complexity assessment in deep learning.

\subsection{Aleatoric Uncertainty and Mixture Models}
To generalize beyond scalar point estimates, probabilistic regression methods explicitly model predictive uncertainty. Heteroscedastic Regression \cite{nix1994estimating} addresses this by modeling \textit{Aleatoric Uncertainty}, the noise inherent in the data itself. Since our goal is to estimate the likelihood of error for a specific input, an intrinsic property of the sample relative to the model, our work is grounded in the principles of heteroscedastic regression.

As formalized in deep learning by Kendall and Gal \cite{kendall2017uncertainties}, these models output both a mean $\mu(x)$ and a variance $\sigma^2(x)$, optimized via the Gaussian Negative Log-Likelihood (NLL). This allows the network to attenuate the loss for ambiguous samples by predicting high variance, effectively learning to express lack of confidence rather than forcing a mean prediction.

To theoretically approximate any conditional probability density function, including multimodal ones, Mixture Density Networks (MDNs) \cite{bishop1994mixture} are often employed. By outputting the parameters of a Gaussian Mixture Model, MDNs can capture multiple modes explicitly. We note that while implicit density estimation via Energy-Based Models (EBMs) \cite{lecun2006tutorial, du2020implicitgenerationgeneralizationenergybased} offers an alternative class of state-of-the-art performance, they typically require computationally expensive sampling (e.g., MCMC) during inference.

However, MDNs are notoriously unstable in practice. Makansi et al. \cite{makansi2019overcoming} highlight that MDNs often suffer from mode collapse and training instabilities. They proposed an Evolving Winner-Takes-All (EWTA) loss to prevent hypothesis collapse during training. Despite these challenges, we compare against the standard MDN to evaluate the trade-off between the theoretical flexibility of mixtures and the stability of our proposed approach.

\subsection{Deep Imbalanced and Non-Parametric Regression}
Our specific focus on bimodal error distributions relates closely to the broader field of Deep Imbalanced Regression (DIR). While standard regression assumes uniform coverage of the target space, real-world data is often highly skewed. Yang et al. \cite{yang2021delving} formally define DIR as learning from data with continuous targets where the distribution is non-uniform, emphasizing that accuracy in rare and extreme values is crucial for generalization.

In the context of deep learning, recent work has attempted to mitigate this imbalance through density estimation and statistical adjustment. Yang et al. \cite{yang2021delving} proposed Label Distribution Smoothing (LDS), which uses kernel density estimation to smooth empirical label frequencies, allowing the model to weight samples based on their effective density. Similarly, Ren et al. \cite{ren2022balanced} introduced Balanced MSE, which statistically adjusts the training objective to compensate for the shift between the training distribution and the test metric.

Other deep learning approaches focus on sample generation or explicit weighting. Tian et al. \cite{tian2023unbalanced} proposed DIRVAE, a generative framework combining Variational Autoencoders with LSTMs to synthesize missing minority samples. Steininger et al. \cite{steininger2021density} introduced DenseWeight, which scales gradients inversely to sample density using Kernel Density Estimation (KDE). While effective, these methods primarily focus on re-weighting or augmenting data to fix the imbalance, rather than modeling the distributional shape of the prediction error itself.

Outside of deep learning, statistical modeling has explicitly addressed bimodality through parametric families. For instance, Vasconcelos et al. \cite{vasconcelos2021new} proposed the Odd Log-Logistic Exponential Gaussian (OLLExGa) for agricultural data. By extending the normal distribution with additional shape parameters, this method achieves the flexibility required to fit two distinct modes. However, applying such parametric models to deep neural networks is challenging, as the latent error distribution of a network is unknown a priori and may not conform to specific rigid functional forms.

To address distribution shape without enforcing strict parametric assumptions, Quantile Regression \cite{koenker2001quantile} offers a robust alternative. Rather than estimating a single mean, this approach minimizes the \textit{Pinball Loss} to learn conditional quantiles. Formally, for a given quantile level $\tau \in (0,1)$, the model learns a value $q$ such that the probability of the target being less than $q$ is $\tau$. By estimating multiple quantiles simultaneously, the model can approximate the bounds and median of complex, multi-modal distributions without assuming normality \cite{meinshausen2006quantile}.

Similarly, Parzen \cite{parzen1962estimation} established foundational work on consistent non-parametric density estimation using kernel functions. In the context of deep learning, Frogner et al. \cite{frogner2015learning} introduced a loss function based on the Wasserstein distance (Earth Mover's Distance) for multi-label learning, demonstrating that penalizing predictions based on semantic distance encourages smoothness and structural alignment. Our work adapts these distributional distances to regression, using them to explicitly align the predicted error distribution with the bimodal ground truth.

\subsection{Application: Complexity and Difficulty Assessment}
Quantifying the difficulty of a training sample is central to paradigms such as Curriculum Learning \cite{bengio2009curriculum}, which posits that training on samples in increasing order of complexity improves convergence and generalization. Hacohen and Weinshall \cite{hacohen2019power} further formalized this by decomposing curriculum learning into difficulty scoring and pacing functions, showing that ideal curricula modify the optimization landscape to be steeper around the global minimum.

While traditional curricula often rely on loss magnitude, recent approaches define difficulty more intrinsically. For instance, Toneva et al.~\cite{tonevaempirical} characterize difficulty via forgetting events, while Baldock et al.~\cite{baldock2021deep} use prediction depth to rank samples from easy to hard. Crucially, however, these methods effectively flatten difficulty into a continuous scalar index. We argue that this assumes a smooth spectrum of hardness that obscures the underlying structure of the problem: difficulty in deep classification is often \textit{structurally bimodal} (distinctly easy vs. distinctly confusing). Standard regression losses like MSE fail in this regime because they are minimized by the \textit{arithmetic mean}, causing predictions to collapse into the sparse region between the two modes rather than identifying the distinct clusters.

While methods such as Deep Ensembles~\cite{lakshminarayanan2017simple} and MC-Dropout~\cite{gal2016dropout} provide robust uncertainty quantification to address this, they typically require multiple forward passes or extensive memory overhead. Our work focuses specifically on \textit{single-pass} complexity assessment, where the goal is to imbue a standard regression model with distributional awareness without the computational cost of ensembling.


\section{Methodology}\label{sec:method}

To address the limitations of existing methods, specifically the tendency of standard regression to \textit{collapse to the mean} and the \textit{optimization instability} of mixture models, we propose a family of distribution-aware loss functions. These losses combine pointwise accuracy with distributional alignment and range matching. Unlike parametric approaches (e.g., OLLExGa \cite{vasconcelos2021new}) that assume rigid distributions, or Mixture Density Networks (MDN) that suffer from stability issues, our approach directly minimizes empirical divergences without parametric assumptions.

Our framework is modular, consisting of three components: a pointwise error metric, a distributional divergence metric, and a range alignment term.

\subsection{Component 1: Pointwise Fidelity}
The first component represents the typical regression objective: minimizing the distance between the prediction and the target value. While our framework accommodates any standard metric (e.g., Mean Absolute Error, Huber Loss), we utilize the Root Mean Squared Error (RMSE) to penalize large deviations.

For numerical stability, we include a small constant $\epsilon$ (e.g., $10^{-8}$) inside the radical to ensure gradients remain bounded when the error approaches zero. Since \textit{D} represents a distance‑like quantity between predictions and targets, we define the raw pointwise metric as:
\begin{equation}
    D_{\text{RMSE}} = \sqrt{\left(\frac{1}{n} \sum_{i=1}^{n} (y_i - \hat{y}_i)^2\right) + \epsilon}
    \label{eq:raw_rmse}
\end{equation}

\subsection{Component 2: Distributional Fidelity}
Standard pointwise losses (like Eq. \ref{eq:raw_rmse}) treat samples independently. Consequently, a model can achieve low MSE by predicting the conditional mean for every sample, even if the underlying problem structure is bimodal. This \textit{collapse to the mean} ignores the aggregate shape of the error distribution, which is the primary signal of interest in complexity assessment.

To recover this structure, we augment the loss with a divergence term $D(F_{\hat{y}}, F_y)$ that penalizes differences between the distribution of predictions $F_{\hat{y}}$ and the distribution of targets $F_y$ within a batch.

A common choice for matching distributions is the Kullback-Leibler (KL) Divergence. However, KL-Divergence requires the estimation of continuous probability densities from discrete batches. In regression settings, this necessitates Kernel Density Estimation (KDE) or histogram binning, which can be sensitive to hyperparameter choice. Furthermore, KL-Divergence is theoretically undefined or infinite when the support of the predicted and target distributions do not overlap, a common occurrence during the early stages of training.

To avoid these issues, we select \textbf{Statistical Distance Metrics} over density-based divergences. Unlike KL, metrics such as the Wasserstein and Cramér distances operate directly on the geometry of the CDFs or quantiles \cite{arjovsky2017wasserstein}. This provides consistent, non-vanishing gradients even when the distributions are disjoint. We investigate two specific metrics:

\begin{enumerate}
    \item \textbf{Cramér Distance:} Defined as the integrated squared difference between the Cumulative Distribution Functions (CDFs):
    \begin{equation}
        D_{\text{Cramér}} = \int_{-\infty}^{\infty} (F_{\hat{y}}(x) - F_y(x))^2 \, dx
        \label{eq:cramer_distance}
    \end{equation}
    
    \item \textbf{Wasserstein Distance:} Also known as the Earth Mover's Distance, defined for 1D distributions as the $L_1$ distance between CDFs \cite{villani2008optimal}:
    \begin{equation}
        D_{\text{Wasserstein}} = \int_{-\infty}^{\infty} |F_{\hat{y}}(x) - F_y(x)| \, dx
        \label{eq:wasserstein_distance}
    \end{equation}
\end{enumerate}

These metrics offer complementary strengths. As noted in \cite{bellemare2017cramer}, the squared term in the Cramér distance makes it highly sensitive to higher-order moment differences, which is beneficial for separating distinct modes. Conversely, the Wasserstein distance provides a linear gradient proportional to the transport cost, excelling at detecting smooth shifts in probability mass.

\subsection{Component 3: Range Alignment}
In bimodal regression, models often shrink the predicted range to minimize variance. To explicitly counter this, we introduce a range alignment penalty:
\begin{equation}
    D_\text{Range} = |\hat{y}_{max} - \hat{y}_{min} - (y_{max} -  y_{min})|
    \label{eq:range_error}
\end{equation}

\subsection{Scale Invariance and Final Loss}
To ensure balanced optimization between error types, we apply a non-linear normalization to map all raw metrics $D$ into a bounded interval $[0, 1]$. The normalization function $\mathcal{N}$ is given by:
\begin{equation}
    \mathcal{N}(D) = 1 - \frac{1}{1 + D}
    \label{eq:normalization}
\end{equation}

We note that minimizing the normalized RMSE component ($\mathcal{N}(D_{\text{RMSE}})$) remains equivalent to minimizing the standard MSE. Since $\mathcal{N}(x)$ and $f(x)=\sqrt{x+\epsilon}$ are both strictly monotonically increasing functions for $x \ge 0$, their composition preserves the global minimizer of the standard squared error.

The final loss function is the weighted sum of the normalized components:
\begin{equation}
    \mathcal{L} = \mathcal{N}(D_{\text{RMSE}}) + \alpha \cdot \mathcal{N}(D_{\text{Distribution}}) + \beta \cdot \mathcal{N}(D_{\text{Range}})
    \label{eq:combined_loss}
\end{equation}
where $D_{\text{Distribution}}$ is either $D_{\text{Cramér}}$ or $D_{\text{Wasserstein}}$. The hyperparameter $\alpha \geq 0$ controls the distributional alignment, and $\beta \geq 0$ controls the range constraint.

\section{Experimental Setup}\label{sec:setup}

\subsection{Datasets}\label{subsec:datasets}
To rigorously validate our hypothesis, we designed a hierarchical experimental protocol using 17 distinct datasets divided into four specific stages, as summarized in Table~\ref{tab:datasets}. This progression allows us to evaluate the loss functions in increasingly complex environments, ranging from controlled synthetic distributions to high-dimensional visual feature spaces.

\begin{table}[ht]
\centering
\caption{Summary of the 17 datasets used across the four-stage experimental protocol.}
\label{tab:datasets}
\begin{tabular}{cllc}
\toprule
\textbf{ID} & \textbf{Dataset} & \textbf{Type} & \textbf{Target Modality} \\
\midrule
\multicolumn{4}{l}{\textit{\textbf{Stage I: Synthetic Proof of Concept}}} \\
1 & Inverse Square & Synthetic Tabular & Conditional Bimodal \\
2 & Two Path & Synthetic Tabular & Conditional Bimodal \\
\midrule
\multicolumn{4}{l}{\textit{\textbf{Stage II: Controlled Separation (Unimodal $\to$ Bimodal)}}} \\
3 & Airfoil & Real-World Tabular & Unimodal* \\
4 & Bike Sharing (Hour) & Real-World Tabular & Unimodal* \\
5 & Concrete Strength & Real-World Tabular & Unimodal* \\
\midrule
\multicolumn{4}{l}{\textit{\textbf{Stage III: Natural Bimodality}}} \\
6 & Houses (OpenML) & Real-World Tabular & Bimodal \\
7 & Protein Structure & Real-World Tabular & Bimodal \\
8 & Energy Efficiency (Heating Load) & Real-World Tabular & Bimodal \\
9 & Energy Efficiency (Cooling Load) & Real-World Tabular & Bimodal \\
10 & Bike Sharing (Day) & Real-World Tabular & Trimodal \\
\midrule
\multicolumn{4}{l}{\textit{\textbf{Stage IV: Image Complexity Assessment}}} \\
11 & CIFAR-10 & Image Error Prediction & Unimodal \\
12 & Fashion-MNIST & Image Error Prediction & Unimodal \\
13 & SVHN & Image Error Prediction & Unimodal \\
14 & CIFAR-100 & Image Error Prediction & Bimodal \\
15 & Oxford Flowers 102 & Image Error Prediction & Bimodal \\
16 & Food-101 & Image Error Prediction & Bimodal \\
17 & Caltech-256 & Image Error Prediction & Bimodal \\
\bottomrule
\end{tabular}
\vspace{0.2cm}
\footnotesize{* Datasets subjected to separation injection parameter $s \in [0, 1]$ to transition from unimodal to bimodal distributions.}
\end{table}

\subsection{Synthetic and Real-World Data (Stages I--III)}
We begin by evaluating the loss functions on tabular benchmarks categorized by their source and complexity structure:

\textbf{Stage I: Synthetic Proof of Concept.}
We utilize two synthetic datasets, \textit{Inverse Square} and \textit{Two Path} (Rows 1--2), explicitly constructed to exhibit strong conditional bimodality. To ensure reproducibility, we define their generation processes as follows:

\begin{itemize}
    \item \textit{Inverse Square:} Targets are sampled uniformly $y \sim \mathcal{U}[-3, 3]$. The informative feature is generated as $x = y^2 + \epsilon$, with Gaussian noise $\epsilon \sim \mathcal{N}(0, 0.5)$. The model must learn the inverse mapping $x \mapsto y \approx \pm\sqrt{x}$, representing two symmetric modes.
    \item \textit{Two Path:} Data is generated from the noisy parametric equations of a circle with radius $r=5$: $x = r\cos(\phi) + \epsilon_x$ and $y = r\sin(\phi) + \epsilon_y$, where $\phi \sim \mathcal{U}[0, 2\pi]$ and $\epsilon \sim \mathcal{N}(0, 0.3)$. For any input $x$, the conditional density $P(y|x)$ contains two modes corresponding to the upper and lower arcs.
\end{itemize}
Both datasets utilize 2-dimensional feature vectors consisting of the generated signal $x$ and an independent Gaussian noise channel to simulate irrelevant features.

\textbf{Stage II: Controlled Separation.}
To study the transition from unimodal to bimodal distributions, we employ three standard regression datasets: Airfoil \cite{brooks1989airfoil}, Bike Sharing (Hour) \cite{fanaee2014event}, and Concrete Strength \cite{concrete_compressive_strength_165} (Rows 3--5). Since these datasets are originally unimodal, we introduce a controllable separation parameter $S \in [0, 1]$ to artificially induce bimodality. 

The transformation mechanism operates by first identifying the natural latent clusters in the target variable $y$ using K-Means ($K=2$). 
Considering $c_{mid}$ as the midpoint between the cluster centroids, we apply a divergent linear transformation:
\begin{equation}
    y' = 
    \begin{cases} 
    y \cdot (1 - S) + y_{min} \cdot S & \text{if } y \le c_{mid} \\
    y \cdot (1 - S) + y_{max} \cdot S & \text{if } y > c_{mid}
    \end{cases}
\end{equation}
where $S=0$ retains the original unimodal distribution, and $S \to 1$ forces the clusters toward the domain boundaries ($y_{min}, y_{max}$), creating a distinct bimodal structure with a clean separation gap.

\textbf{Stage III: Natural Bimodality.}
We include five real-world datasets (Rows 6--10) that exhibit naturally occurring bimodality without modification. These datasets consist of \textit{California Housing}~\cite{pace1997sparse, california_housing_1997}, \textit{Protein Structure}, Energy Efficiency \cite{energy_efficiency_242} and Bike Sharing (Day) \cite{fanaee2014event} where unobserved latent variables often create distinct regimes in the target variable, making them ideal benchmarks for distribution-aware losses.

\subsection{Image Data Generation (Stage IV)}
A critical contribution of this work is evaluating complexity assessment on high-dimensional visual data. 
For the image complexity assessment, we utilize seven established computer vision datasets: CIFAR-10~\cite{krizhevsky2009learning}, Fashion-MNIST~\cite{xiao2017fashion}, Street View House Numbers (SVHN)~\cite{netzer2011reading}, CIFAR-100~\cite{krizhevsky2009learning}, Oxford Flowers~102~\cite{nilsback2008automated}, Food-101~\cite{bossard2014food}, and Caltech-256~\cite{griffin2007caltech}. Unlike tabular data where targets are intrinsic, the regression targets for image datasets must be derived from a classification task. We utilized a two-step procedure for the image benchmarks (Rows 11--17):

\textbf{Target Generation (Classification).} We trained an EfficientNet-V2-M classifier \cite{tan2019efficientnet} on the training split of each dataset. The model was optimized using Adam with a weight decay of $10^{-4}$. After convergence, we utilized the unseen test split to calculate the scalar error $y_i = 1 - P(c_i|x_i)$. This scalar serves as the ground truth complexity ($y$) for the subsequent regression task. We utilize this bounded metric $[0,1]$ rather than Cross-Entropy ($-\log P$) to ensure numerical stability.
    
\textbf{Feature Extraction.} To isolate the performance of the loss function from the capacity of the feature extractor, we froze the backbone of an EfficientNet-B0 pre-trained on ImageNet. We replaced the classification head with an identity mapping to extract a 1280-dimensional embedding vector for every image.

Thus, for the regression experiments, the model receives a fixed 1280-dimensional feature vector ($x$) and must predict the scalar classification error ($y$) derived from the test samples in previous step.

\subsection{Data Preprocessing}
\label{subsec:preprocessing}
To ensure training stability and comparability across datasets with varying physical units and magnitudes, we applied standard scaling ($z$-score normalization) to both the input features and the regression targets. This transformation standardizes the data by subtracting the arithmetic mean and dividing by the standard deviation. We note that, to prevent data leakage, these statistics are derived exclusively from the training split and subsequently applied to the validation and test sets. This preprocessing strategy was applied uniformly across all dataset types. For tabular datasets, all numerical input features were standardized. For the image error datasets, the same standardization was applied to the 1280-dimensional feature embeddings before they were fed into the MLP. Finally, all regression targets (e.g., prediction error, physical loads) were standardized to zero mean and unit variance.

\subsection{Proposed Variants and Ablation}
To isolate the individual contributions of the distance metrics and the range constraint, we evaluate four distinct configurations of our framework. This decomposition allows us to address \textbf{RQ2 (Stability)} by comparing the gradient behavior of Cramér vs. Wasserstein, and \textbf{RQ1 (Fidelity)} by ablating the range term.

For brevity, we abbreviate Wasserstein as \textit{Wasser} in the reported results:

\begin{itemize}
    \item \textbf{Cramér-Simple / Wasser-Simple} ($\beta = 0$):
    These variants focus solely on the trade-off between pointwise error and distributional matching. By setting $\beta=0$, we evaluate the capability of the raw distance metrics to recover bimodal structures without explicit boundary constraints.

    \item \textbf{Cramér-Range / Wasser-Range} ($\beta = \alpha / 2$):
    These variants incorporate the range alignment term. We fix $\beta = \alpha/2$ as a heuristic stability constraint. This ensures that the distributional fidelity term ($\alpha$) remains the primary optimization objective, while the range term acts as a soft regularizer to prevent support collapse.
\end{itemize}

\subsection{Network Architecture}\label{subsec:architecture}
To ensure that performance variations are strictly attributable to the proposed loss functions rather than architectural discrepancies, we employ a unified experimental framework across all datasets.

For all 17 datasets, ranging from 2D synthetic data to 1280D image embeddings, we utilize an identical Multi-Layer Perceptron (MLP) architecture. This acts as a consistent probe to evaluate the loss functions. 

The architecture consists of an input layer adapting to the feature dimension (1 for synthetic, 1280 for images), followed by three hidden layers with 512, 256, and 128 units, respectively, and a final bottleneck layer of 64 units before the single scalar output. To mitigate overfitting, we employ Batch Normalization and GELU activation after each hidden layer, along with Dropout rates of 0.3 for the initial layers and 0.2 for deeper layers. All models are implemented in PyTorch \cite{NEURIPS2019_bdbca288} and trained using the Adam optimizer \cite{kingma2014adam} with a learning rate of $10^{-3}$ and a batch size of 64 for 50 epochs. 

For the MDN baseline comparison, we employ a network with a shared hidden representation comprising two fully connected layers with Tanh activations. This shared trunk branches into three separate heads to predict the mixing coefficients ($\pi$), means ($\mu$), and standard deviations ($\sigma$) for $K=5$ Gaussian components. The standard deviations are constrained to be positive using the ELU activation plus a stability constant ($1 + \epsilon$).

\subsection{Baselines}
To rigorously evaluate the proposed framework, we compare against four distinct baselines representing the spectrum from naive deterministic regression to advanced probabilistic modeling. All distinct architectures share a common Multi-Layer Perceptron (MLP) backbone to ensure fair comparison of the loss functions.

\begin{itemize}
    \item \textbf{MDN (NLL) [Primary Competitor]:} 
    A Mixture Density Network \cite{bishop1994mixture} that outputs parameters ($\pi, \mu, \sigma$) for a Gaussian Mixture Model (GMM). Optimized via Negative Log-Likelihood (NLL), this is the theoretical gold standard for multimodal regression but is often prone to optimization instability.
    
    \item \textbf{HMLP (GaussianNLL) [Probabilistic Baseline]:} 
    A Heteroscedastic MLP that outputs both a mean $\mu(x)$ and a variance $\sigma^2(x)$. It is trained using the Gaussian Negative Log-Likelihood (GaussianNLL) loss. This represents the standard approach for modeling aleatoric uncertainty under a unimodal Gaussian assumption.
    
    \item \textbf{MLPQ (Quantile) [Robust Non-Parametric Baseline]:} 
    A Quantile Regression MLP that outputs fixed quantiles (e.g., $\tau \in \{0.1, 0.5, 0.9\}$) instead of distribution parameters. Optimized via the Pinball Loss (QuantileLoss), it provides a distribution-free estimate of prediction intervals and serves as a robust non-parametric competitor.
    
    \item \textbf{MLP (MSE) [Naive Baseline]:} 
    A standard MLP optimized with Mean Squared Error (MSE). This serves as the deterministic lower bound, predicting the conditional mean $E[y|x]$. It uses the same architecture as our proposed method but lacks distributional awareness.
\end{itemize}

\subsection{Evaluation Metrics}\label{subsec:metrics}
To comprehensively assess the performance, we employ a dual-metric strategy that evaluates both sample-level accuracy and distributional alignment.

We utilize \textbf{Root Mean Squared Error (RMSE)} and \textbf{Mean Absolute Error (MAE)} to measure the precision of individual predictions. While our primary goal is distributional fidelity, these metrics serve as a control to ensure that the model retains predictive power and does not simply output random samples from a correct distribution.

Since standard regression metrics fail to penalize mode collapse, we introduce three distribution-aware metrics:

\begin{enumerate}
    \item \textbf{Jensen-Shannon (JS) Divergence:} To quantify the overlap between the predicted and ground-truth error distributions, we utilize the JS Divergence. Unlike KL-Divergence, JS is symmetric and bounded in $[0, 1]$, providing a stable metric for comparing the probability distributions derived from the scalar predictions.

    \item \textbf{Wasserstein Distance:} While our training objective incorporates a learned approximation of the Wasserstein distance, we explicitly calculate the exact distance (Earth Mover's Distance) during evaluation. This serves two purposes: first, it acts as an objective, critic-independent verification of convergence, ensuring that the generator is not merely fooling a sub-optimal critic. Second, unlike JS divergence, which relies on density overlap, $W_1$ captures the geometry of the error space, quantifying the physical "effort" required to transport the predicted distribution to the ground truth.

    \item \textbf{Structural Fidelity ($\Delta_{BC}$):} To explicitly evaluate whether the model captures the structural complexity of the error surface (e.g., the bimodal nature of \textit{Easy} vs. \textit{Hard} samples), we first compute the Bimodality Coefficient (BC)~\cite{pfister2013good} of the predictions:
    
\begin{equation}
BC = \frac{\gamma^2 + 1}{\kappa + \frac{3(n-1)^2}{(n-2)(n-3)}}
\end{equation}
where $\gamma$ is skewness, $\kappa$ is excess kurtosis, and $n$ is sample size. Values $>0.555$ indicate bimodality ~\cite{pfister2013good,freeman2013assessing, kang2019development}.
We define \textit{Structural Fidelity} as the absolute deviation of the predicted BC from the target:
\begin{equation}
\Delta_{BC} = |BC_{\text{target}} - BC_{\text{pred}}|
\end{equation}
A $\Delta_{BC}$ close to 0 indicates that the model successfully recovers the distribution structure, whereas high values quantify the severity of the regression to the mean (unimodal collapse).
\end{enumerate}

\section{Results}\label{sec:results}
We structure our analysis hierarchically, moving from controlled synthetic environments to the motivating problem of deep learning complexity assessment. We analyze performance across four distinct dataset types: Synthetic, Unimodal Real-World Tabular, Bimodal Real-World Tabular, and Image Error Prediction. There is also one Trimodal Real-World Tabular for comparison.

\subsection{Stage I: Validation on Synthetic Distributions}
\label{subsec:res_synthetic}

To verify if the proposed loss functions can successfully capture non-Gaussian, bimodal structures in a controlled environment, and to quantify the trade-off between pointwise 
fidelity 
(RMSE) and distributional fidelity (JS Divergence). We evaluate the models on the \textit{Inverse Square} and \textit{Two Path} synthetic datasets. 

\textbf{Stability and Pointwise Fidelity:}
The primary challenge in bimodal regression is maintaining the training stability of deterministic models while introducing probabilistic flexibility. As shown in Table \ref{tab:stage1_results}, the \textit{Wasser-Simple} configuration achieves an RMSE of 2.62, exactly matching the HMLP baseline (2.62) and outperforming the standard MSE baseline (2.70). This confirms that our proposed loss preserves the robust convergence properties of the MLP backbone. In contrast, the MDN baseline, while theoretically capable of modeling mixtures, suffers from severe optimization instability, resulting in a significantly degraded RMSE of 3.82.

\textbf{Distributional Trade-off Analysis:}
Beyond pointwise accuracy, we analyze the trade-off between stability and distributional fidelity using Jensen-Shannon (JS) Divergence. 

\textit{Improving Fidelity while maintaining Stability:} The HMLP baseline, constrained by its unimodal Gaussian assumption, yields a high JS Divergence of 0.73. By incorporating our transport-based loss, the \textit{Wasser-Simple} model reduces this divergence to 0.63 without any penalty to RMSE (remaining at 2.62). This represents an improvement in distributional alignment.
    
\textit{Maximizing Fidelity:} For applications prioritizing pure distribution matching, the \textit{Cramer-Range} variation reduces JS Divergence further to 0.54, effectively matching the best probabilistic baseline (MDN, 0.54). However, unlike the MDN which sacrifices stability (RMSE 3.82), the \textit{Cramer-Range} maintains a competitive RMSE of 2.94. Additionally, among the tuned configurations, \textit{Cramer-Range} achieves the minimum Wasserstein distance of 1.38.

\textit{Shape Recovery:} In terms of capturing the specific bimodal shape, the \textit{Cramer-Simple} variation excels, achieving a $\Delta_{BC}$ of 0.14, marginally outperforming the MDN (0.15) and significantly improving upon the unimodal baselines (HMLP 0.24, MSE 0.34).

\begin{figure}[ht]
    \centering
    \begin{subfigure}[b]{0.45\textwidth}
        \centering
        \includegraphics[width=\textwidth]{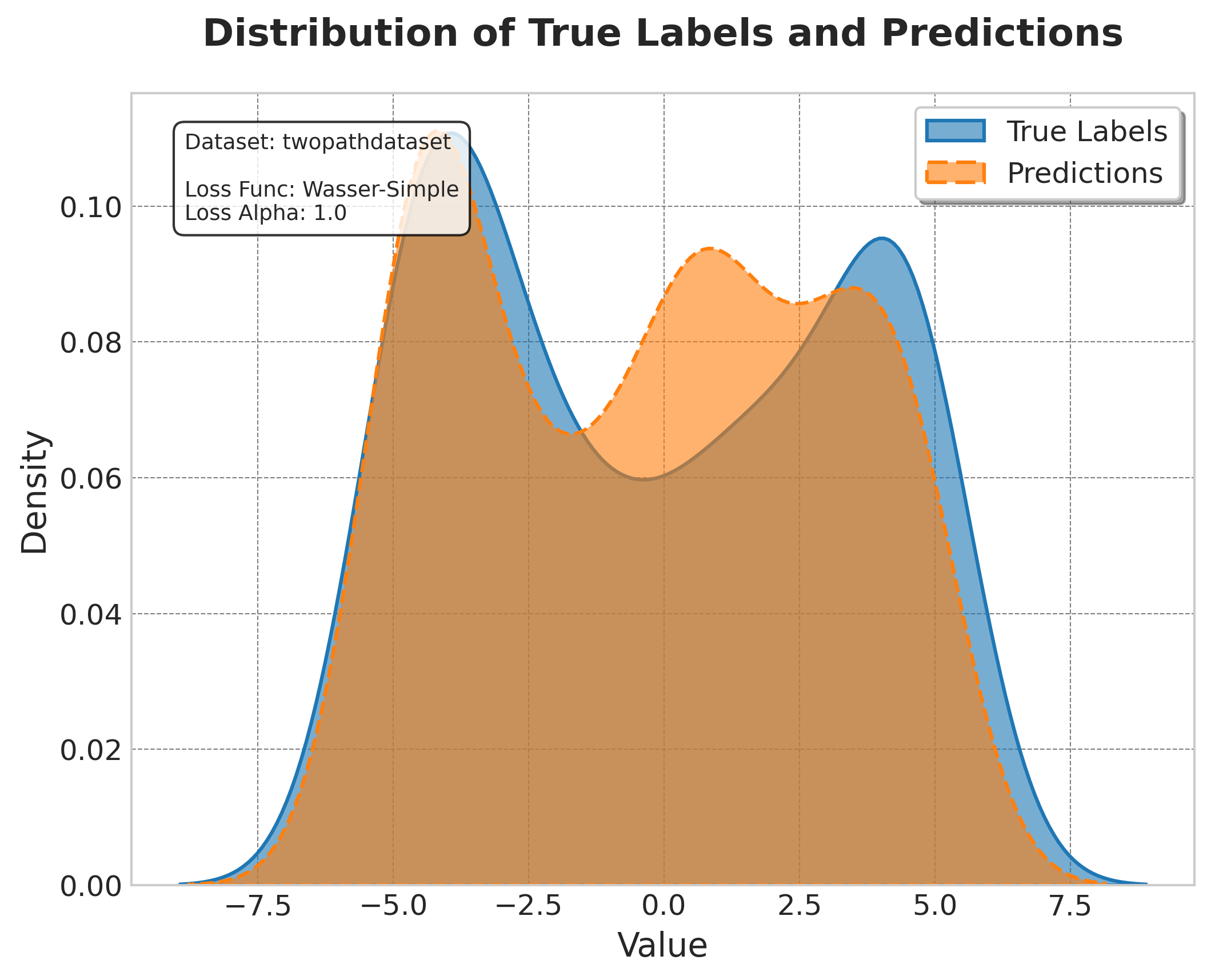} 
        \caption{Proposed ($\alpha=1$)}
    \end{subfigure}
    \hfill
    \begin{subfigure}[b]{0.45\textwidth}
        \centering
        \includegraphics[width=\textwidth]{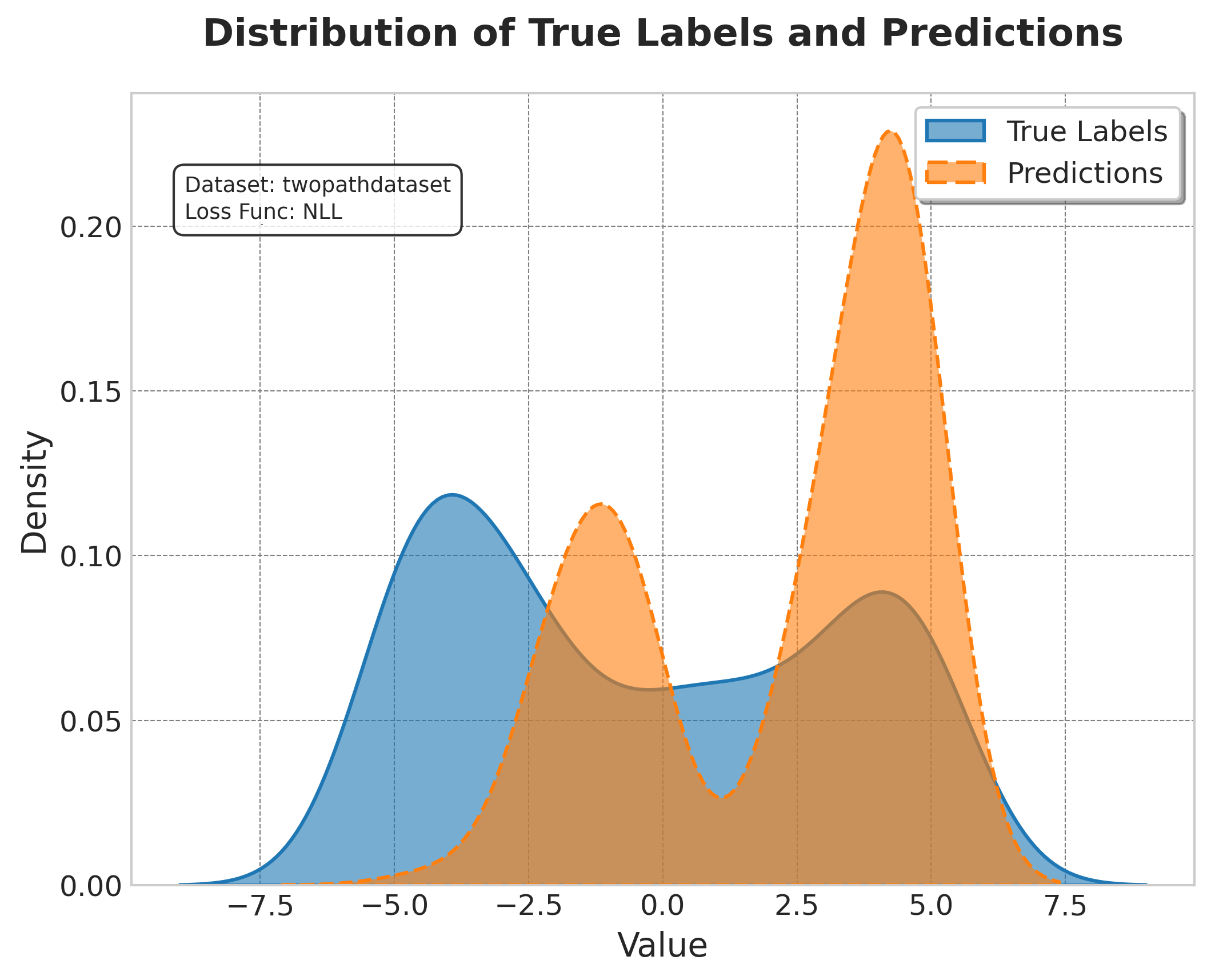} 
        \caption{MDN}
    \end{subfigure}

    \vspace{0.5cm} 

    \begin{subfigure}[b]{0.45\textwidth}
        \centering
        \includegraphics[width=\textwidth]{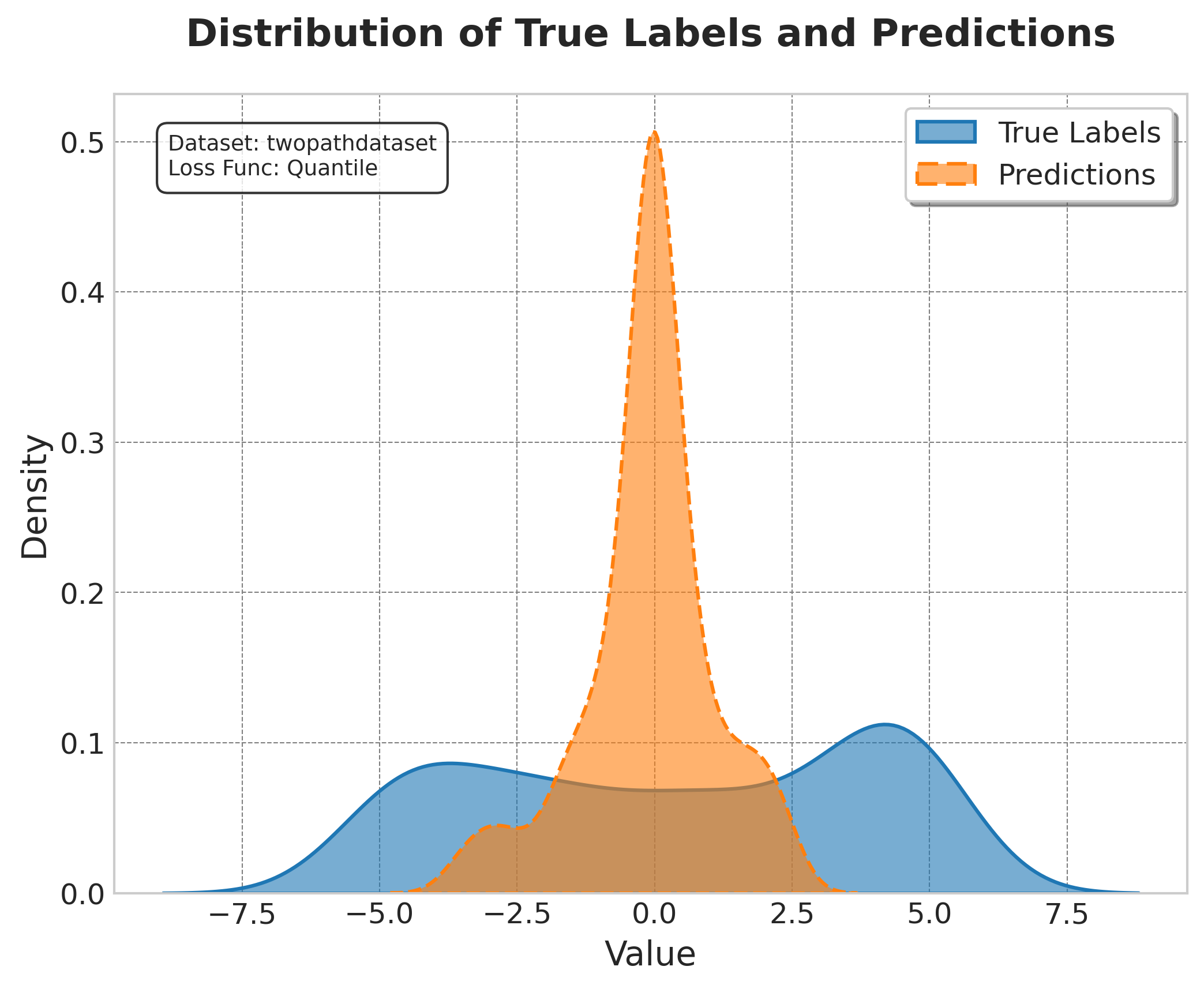} 
        \caption{Quantile Regression}
    \end{subfigure}
    \hfill
    \begin{subfigure}[b]{0.45\textwidth}
        \centering
        \includegraphics[width=\textwidth]{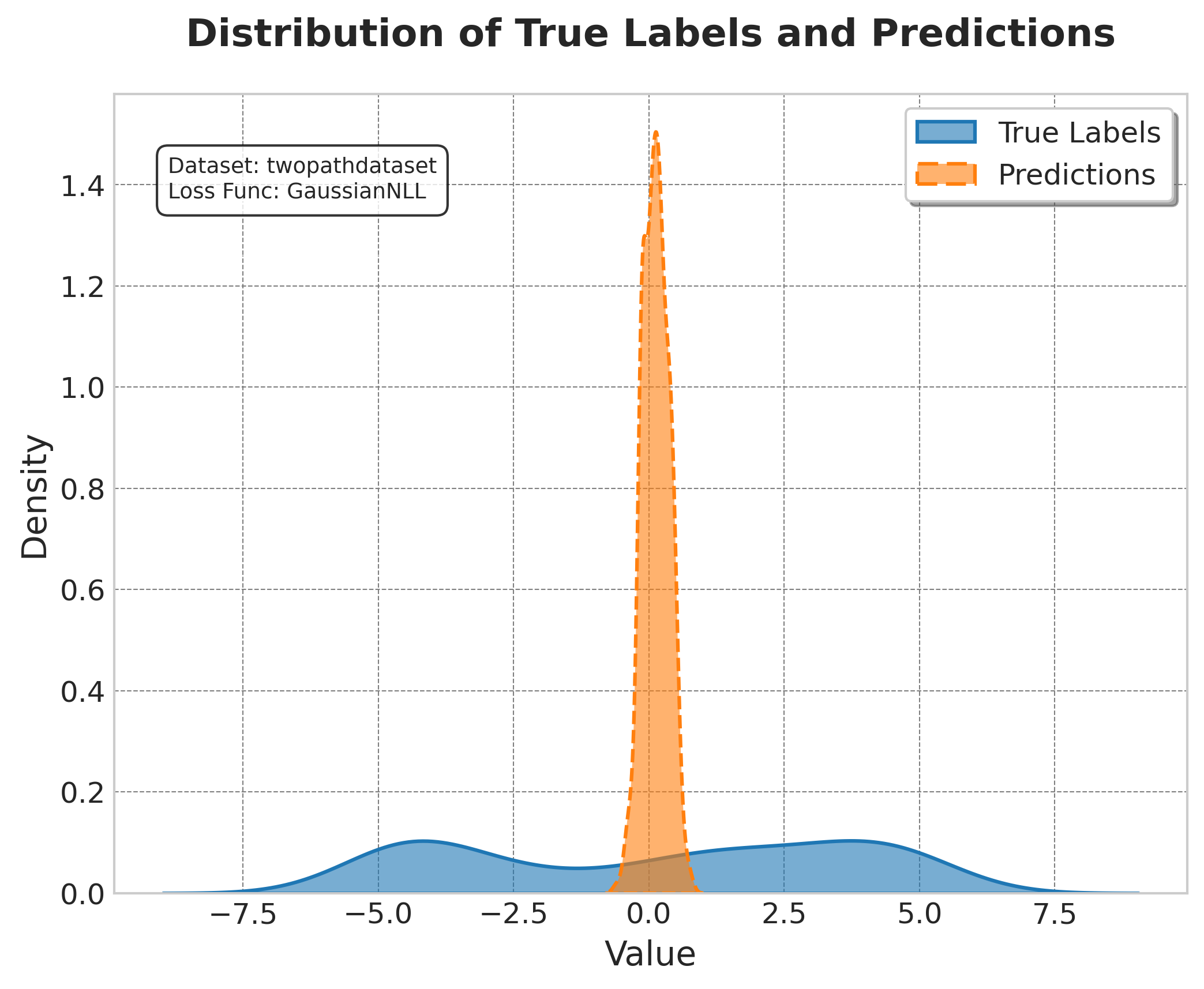} 
        \caption{Heteroscedastic MLP (HMLP)}
    \end{subfigure}

    \caption{Qualitative comparison of predictive distributions (Orange) versus Ground Truth (Blue) on the \textit{Two Path} dataset.}
    \label{fig:stage1_kde_qualitative}
\end{figure}

\textit{Comparison with Default Configuration:} We also evaluate the \textit{Wasser-Simple (Def)} configuration ($\alpha=1$), representing an unweighted, \textit{out-of-the-box} setting. This model highlights a critical limitation of pointwise metrics in bimodal regression: while baselines like HMLP and Quantile achieve lower RMSE, they do so by collapsing to a unimodal average. In contrast, the default configuration preserves the bimodal geometry, achieving exceptional fidelity ($\Delta_{BC}$ of 0.03, Wasserstein of 0.30, JS Divergence of 0.37). It drastically outperforms the MDN ($\Delta_{BC}$ of 0.15), which, despite being probabilistic, suffers from optimization instability, manifesting as shifted modes and exaggerated peak amplitudes (as observed in the \textit{Two Path} dataset in Figure \ref{fig:stage1_kde_qualitative}), demonstrating that our method provides superior structural recovery even without hyperparameter tuning.

\textbf{Qualitative Shape Recovery:}
To visualize the capacity for mode recovery, Figure \ref{fig:stage1_kde_qualitative} presents the prediction densities on the \textit{Two Path} dataset. Visual inspection reveals that while the HMLP and Quantile baselines suffer from unimodal collapse, the MDN recovers the bimodal structure but exhibits optimization instability, evidenced by shifted modes and exaggerated peak amplitudes. In contrast, the proposed method accurately splits the probability mass, aligning closely with the ground truth geometry.

In summary, the Stage I results confirm that the proposed framework successfully bridges the gap between the stability of deterministic baselines and the expressiveness of mixture models. Specifically, \textit{Wasser-Simple} offers the stability of HMLP with improved divergence metrics, while \textit{Cramer-Range} matches the distributional fidelity of MDNs with significantly better stability. Overall, the proposed methods achieve a balance between stability (RMSE) and shape recovery (Wasserstein/$\Delta_{BC}$).

Table \ref{tab:stage1_results} summarizes the performance of the proposed loss functions (averaged across both datasets) against the baselines.

\begin{table}[ht]
\centering
\caption{Performance comparison on synthetic datasets.}
\label{tab:stage1_results}
\footnotesize 
\setlength{\tabcolsep}{3pt} 
\begin{tabular}{lccccc}
\toprule
\textbf{Configuration} & \textbf{Test Loss} & \textbf{RMSE} $\downarrow$ & \textbf{Wasserstein} $\downarrow$  &\textbf{JS Div} $\downarrow$ 
& \textbf{$\Delta_{BC}$} $\downarrow$ \\
\midrule
\multicolumn{6}{l}{
} \\
Wasser-Simple & 0.535 & \textbf{2.62} & 1.80  &0.63 
& 0.29 \\
Wasser-Range  & 0.581 & 2.90 & 1.71  &0.59 
& 0.34 \\
Cramer-Simple & 0.542 & 2.77 & 1.73  &0.62 
& 0.14 \\
Cramer-Range  & 0.576 & 2.94 & 1.38  &0.54 
& 0.39 \\
Wasser-Simple (Def)  & 0.777 & 3.65 & \textbf{0.30}  &\textbf{0.37} 
& \textbf{0.03} \\
\midrule 
\multicolumn{6}{l}{
} \\
MDN (NLL)     & \textbf{0.071} & 3.82 & 1.61  &0.54 
& 0.15 \\
HMLP (Gauss)  & 0.280 & 2.62 & 2.15  &0.73 
& 0.24 \\
MLPQ (Quant)  & 0.214 & 2.87 & 1.56  &0.55 
& 0.31 \\
MSE (Standard)& 1.050 & 2.70 & 2.20  &0.72 & 0.34 \\
\bottomrule
\end{tabular}
\end{table}

\subsection{Stage II: Continuous Mode Separation Analysis}
\label{subsec:res_stage2}

To evaluate the stability of the loss landscape during the critical phase transition from unimodal to bimodal distributions. We utilize a controlled 
separation parameter $S \in [0, 1]$ applied to the synthetic datasets. The bimodal targets are generated via a split-and-shift mechanism: data points below the K-Means midpoint are linearly interpolated towards the domain minimum, while those above are shifted towards the maximum, governed by $S$. At $S=0$, the distribution is strictly unimodal; as $S \to 1$, the modes diverge into a clear bimodal structure.

Table \ref{tab:stage2_snapshots} presents the performance metrics at critical transition points, addressing \textbf{RQ1} regarding structural recovery and \textbf{RQ2} regarding optimization stability during distribution shifts.

\begin{table}[ht]
\centering
\caption{Performance snapshots at critical separation intervals ($S$) in synthetic datasets.}
\label{tab:stage2_snapshots}
\footnotesize 
\setlength{\tabcolsep}{2.5pt} 
\begin{tabular}{lccccccc}
\toprule
& \multicolumn{2}{c}{\textbf{Unimodal ($S=0$)}} & \multicolumn{2}{c}{\textbf{Transition ($S=0.5$)}} & \multicolumn{3}{c}{\textbf{Bimodal ($S=1$)}} \\
\cmidrule(lr){2-3} \cmidrule(lr){4-5} \cmidrule(lr){6-8}
\textbf{Configuration} & \textbf{RMSE} & \textbf{JS Div} & \textbf{RMSE} & \textbf{JS Div} & \textbf{RMSE} & \textbf{JS Div} & \textbf{$\Delta_{BC}$} \\
\midrule
\textit{Proposed Methods} & & & & & & & \\
Wasser-Simple & 0.939 & 0.427 & 0.467 & 0.387 & 0.509 & 0.441 & 0.105 \\
Wasser-Range & 0.977 & 0.430 & 0.491 & 0.366 & 0.532 & 0.436 & 0.171 \\
Cramer-Simple & 0.951 & 0.457 & 0.481 & 0.392 & 0.501 & 0.472 & 0.129 \\
Cramer-Range & 0.988 & 0.440 & 0.480 & 0.378 & 0.518 & 0.441 & 0.156 \\
Wasser-Simple (Def) & 1.093 & \textbf{0.284} & 0.493 & \textbf{0.322} & 0.581 & \textbf{0.339} & \textbf{0.052} \\
\midrule
\textit{Baselines} & & & & & & & \\
MDN & 1.154 & 0.314 & 0.591 & 0.368 & 0.607 & 0.411 & 0.071 \\
HMLP & 0.968 & 0.403 & 0.472 & 0.404 & 0.535 & 0.525 & 0.143 \\
MLPQ & 0.967 & 0.369 & \textbf{0.460} & 0.393 & 0.503 & 0.469 & 0.061 \\
MSE & \textbf{0.942} & 0.468 & 0.469 & 0.409 & \textbf{0.496} & 0.483 & 0.140 \\

\bottomrule
\end{tabular}
\end{table}

The results highlight distinct behavioral classes among the distribution-aware losses:

\begin{itemize}
    \item \textbf{The Singularity Phenomenon ($S=0 \to 0.1$):} A critical transition occurs at the onset of separation. At $S=0$, the distribution is continuous and unimodal. However, the generation mechanism applies a hard split at the distribution midpoint, linearly interpolating data toward the domain boundaries as $S$ increases. The sharp drop in error metrics at $S=0.1$ marks the singularity where this split creates two statistically resolvable attractors. While Cramer and Wasserstein models adapt immediately to this structural break, the MDN exhibits instability at $S=0$, failing to collapse to the single mean when no gap exists.

    \item \textbf{Cramer vs. Wasserstein (Structure):} While Cramer loss is theoretically a distance metric like Wasserstein, the results show it behaves closer to MSE in practice. At $S=1$, \textit{Cramer-Simple} yields a $\Delta_{BC}$ of 0.129 and \textit{Cramer-Range} 0.156, both comparable to the collapsed HMLP (0.143). This indicates that Cramer loss struggles to drive the gradient towards explicit mode separation. In contrast, \textit{Wasser-Simple (Def)} achieves a $\Delta_{BC}$ of 0.052, confirming superior geometric recovery.
    
    \item \textbf{Information Fidelity (JS Div):} The \textit{Wasser-Simple (Def)} is the only configuration that maintains high distributional fidelity across the entire spectrum, achieving the lowest JS Divergence at both the unimodal singularity (0.284) and the bimodal peak (0.339).
\end{itemize}

\begin{figure}[ht]
    \centering
    \begin{subfigure}[b]{0.32\textwidth}
        \includegraphics[width=\textwidth]{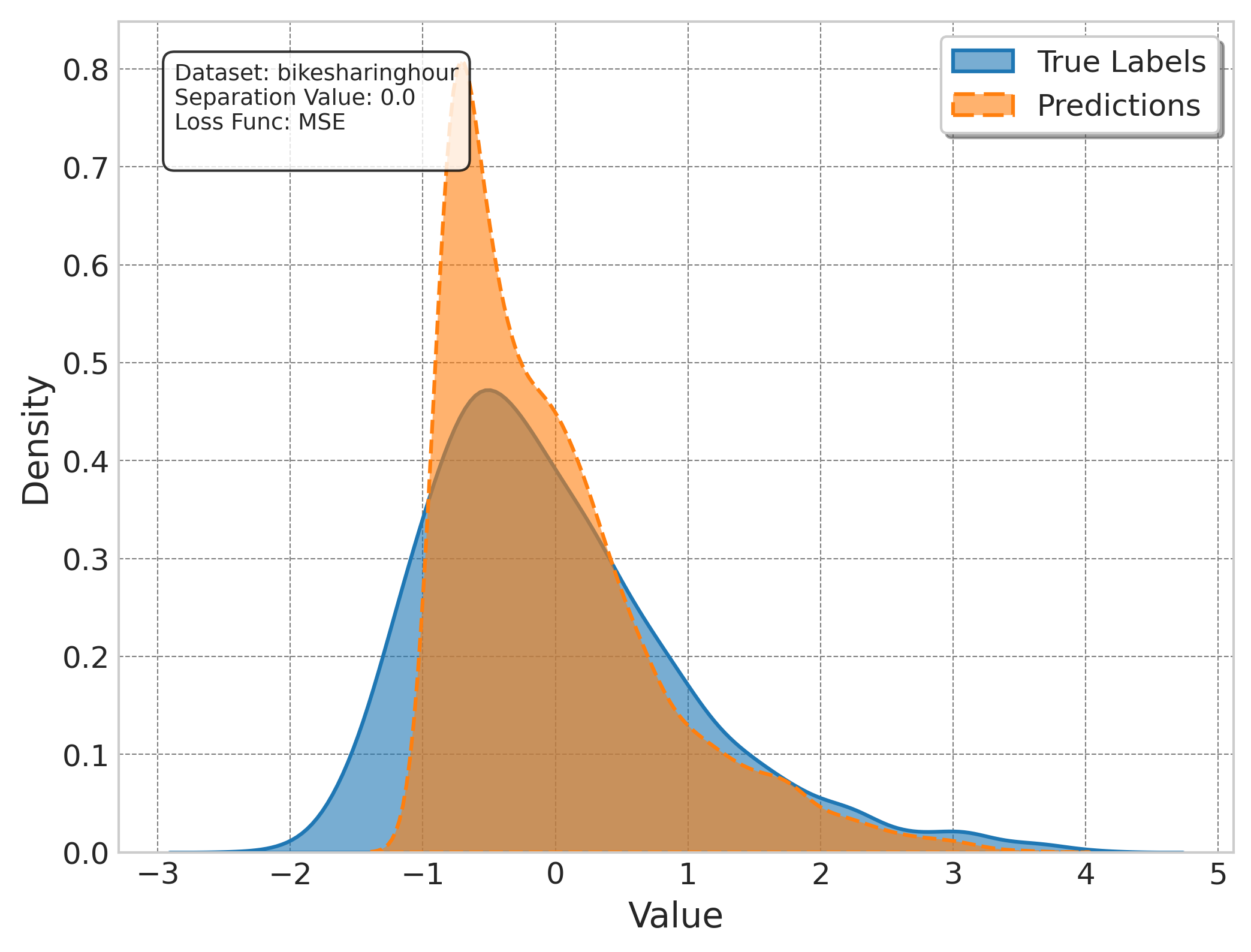}
        \caption{MSE ($S=0.0$)}
    \end{subfigure}
    \hfill
    \begin{subfigure}[b]{0.32\textwidth}
        \includegraphics[width=\textwidth]{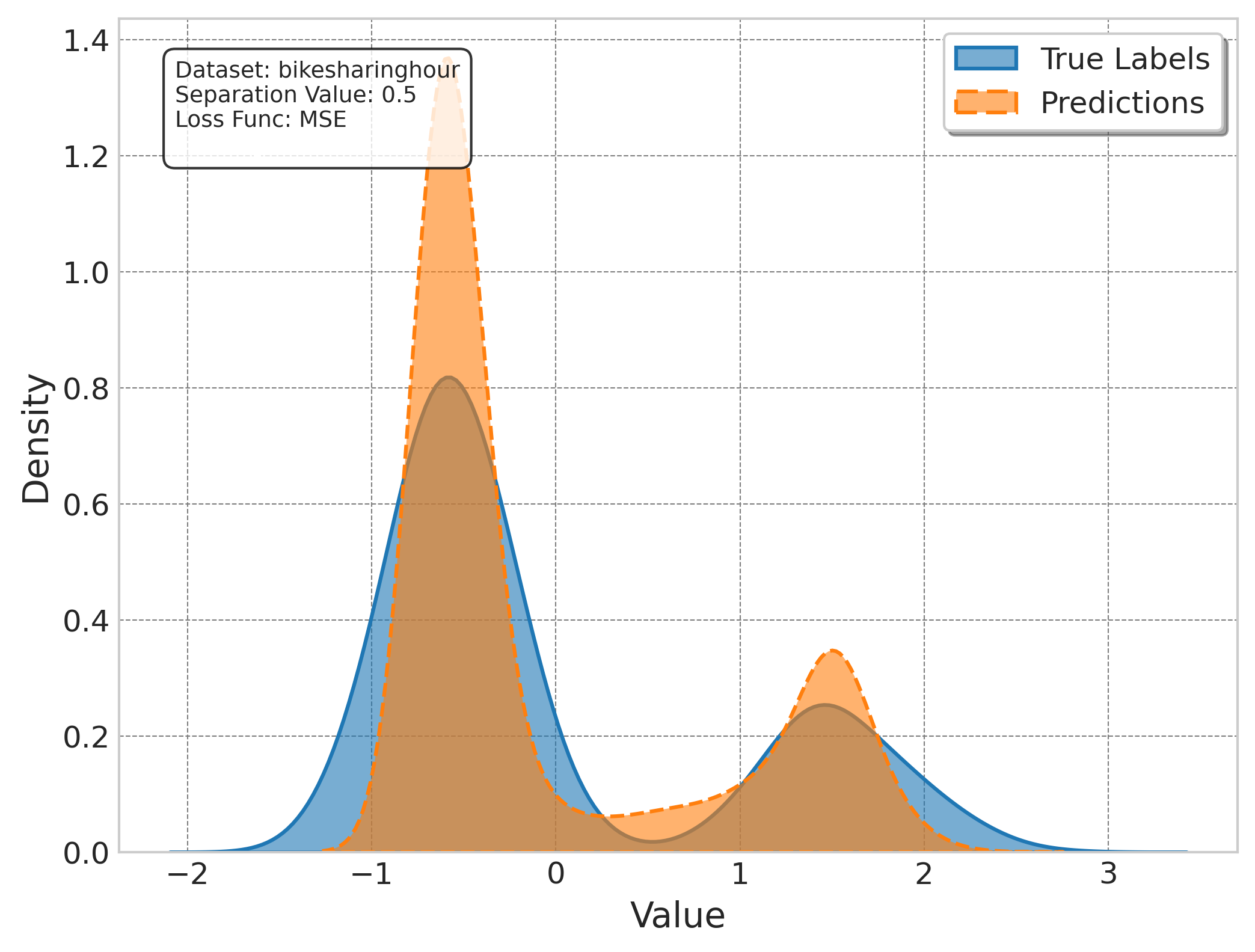}
        \caption{MSE ($S=0.5$)}
    \end{subfigure}
    \hfill
    \begin{subfigure}[b]{0.32\textwidth}
        \includegraphics[width=\textwidth]{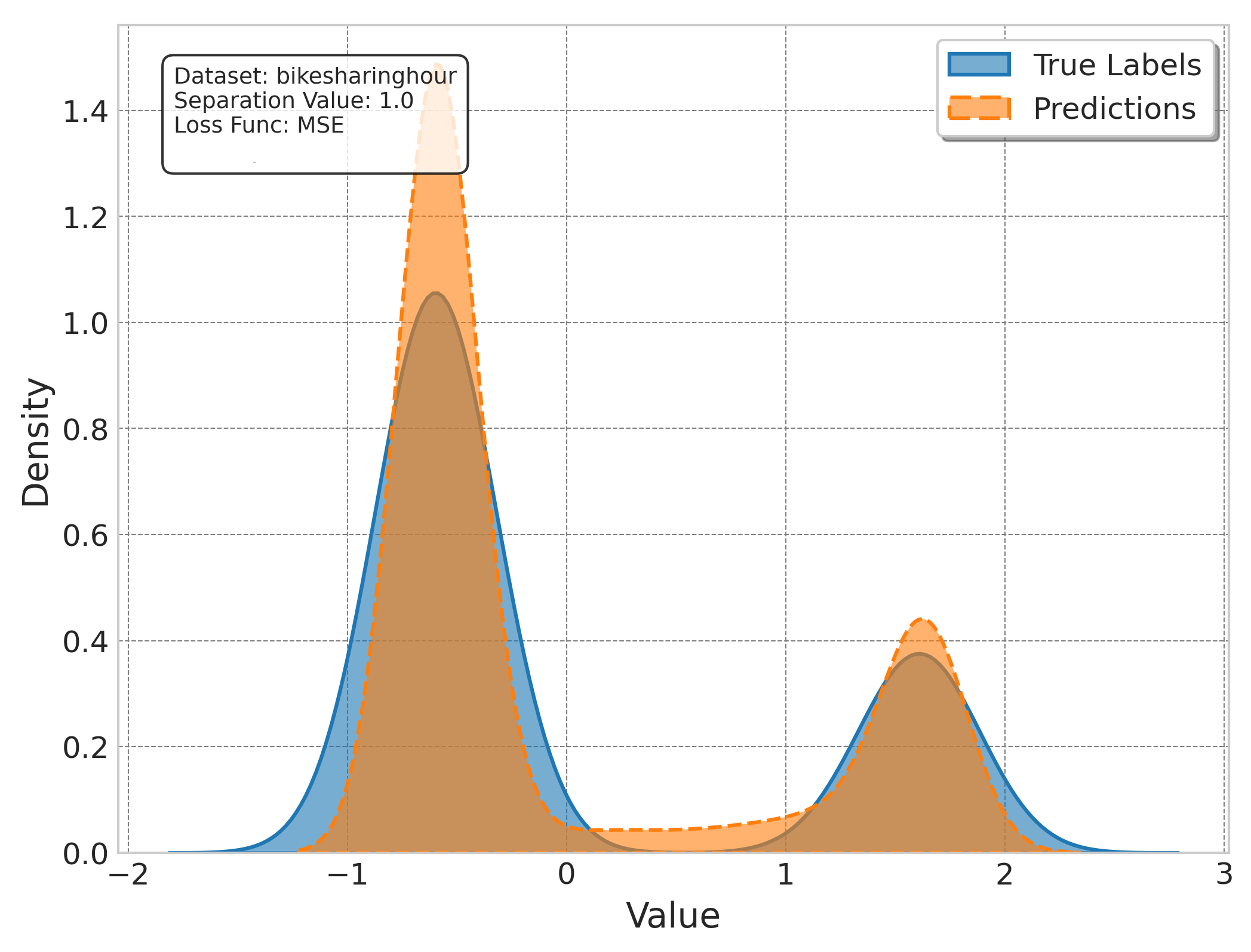}
        \caption{MSE ($S=1.0$)}
    \end{subfigure}
    
    \vspace{0.1cm}
    
    \begin{subfigure}[b]{0.32\textwidth}
        \includegraphics[width=\textwidth]{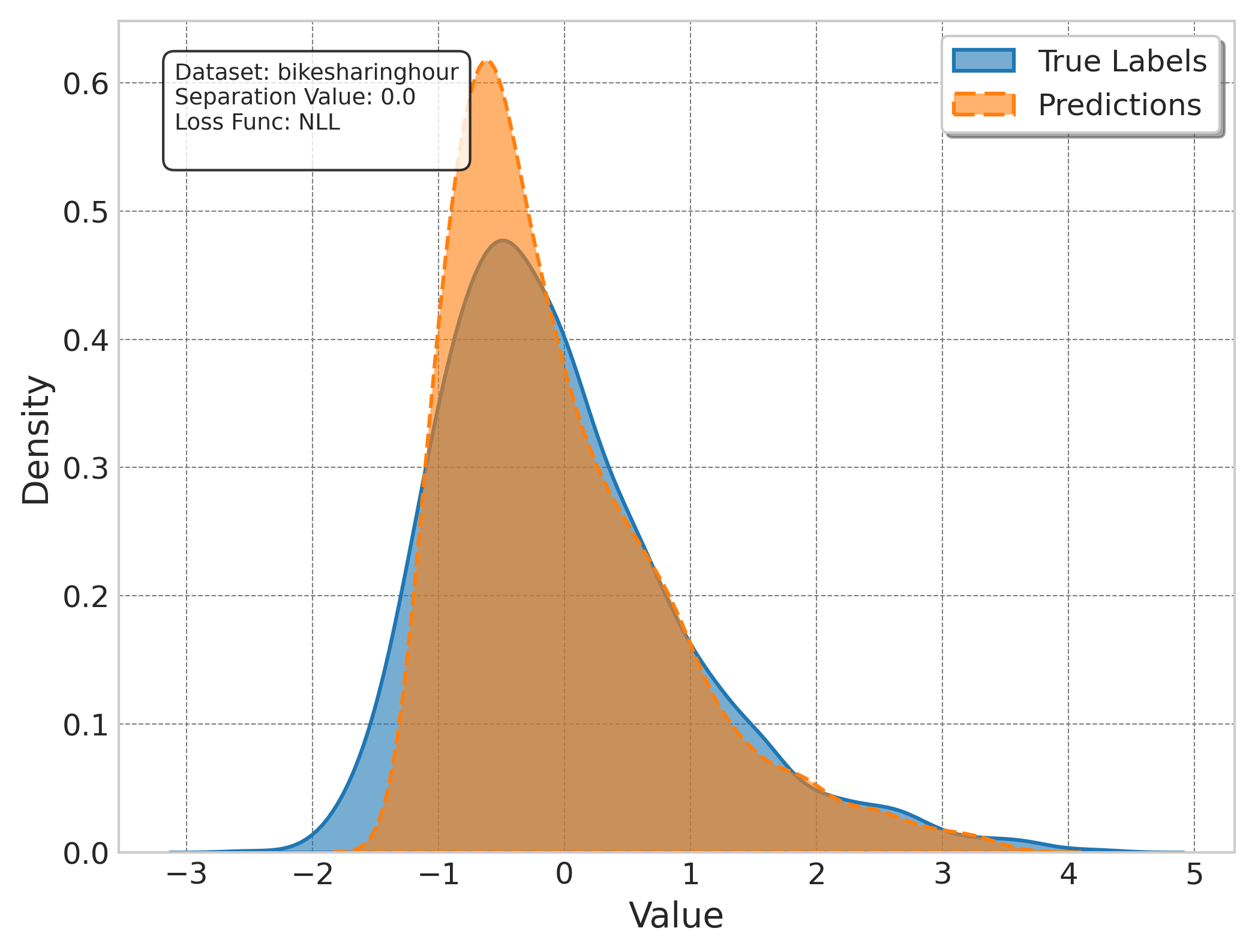}
        \caption{MDN ($S=0.0$)}
    \end{subfigure}
    \hfill
    \begin{subfigure}[b]{0.32\textwidth}
        \includegraphics[width=\textwidth]{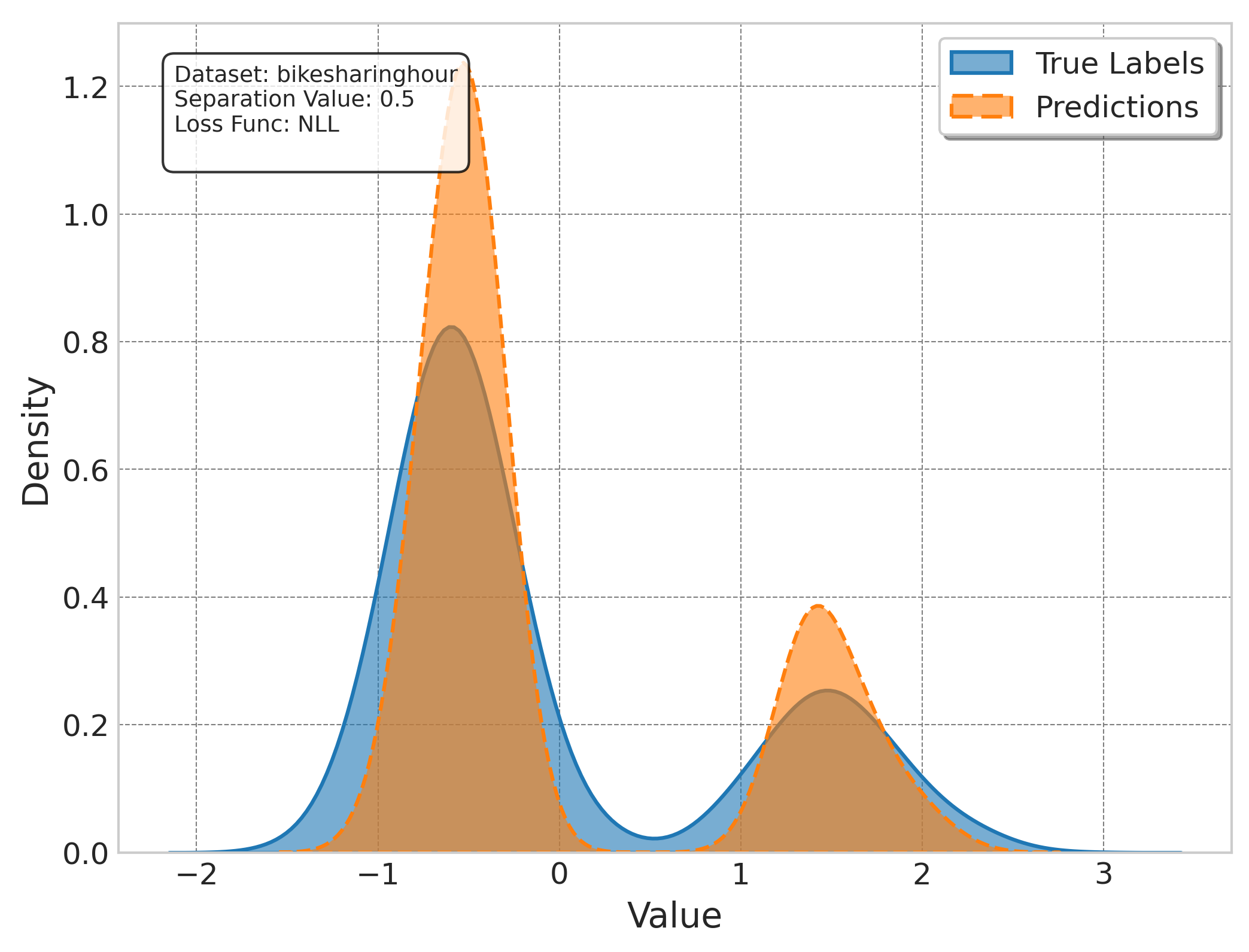}
        \caption{MDN ($S=0.5$)}
    \end{subfigure}
    \hfill
    \begin{subfigure}[b]{0.32\textwidth}
        \includegraphics[width=\textwidth]{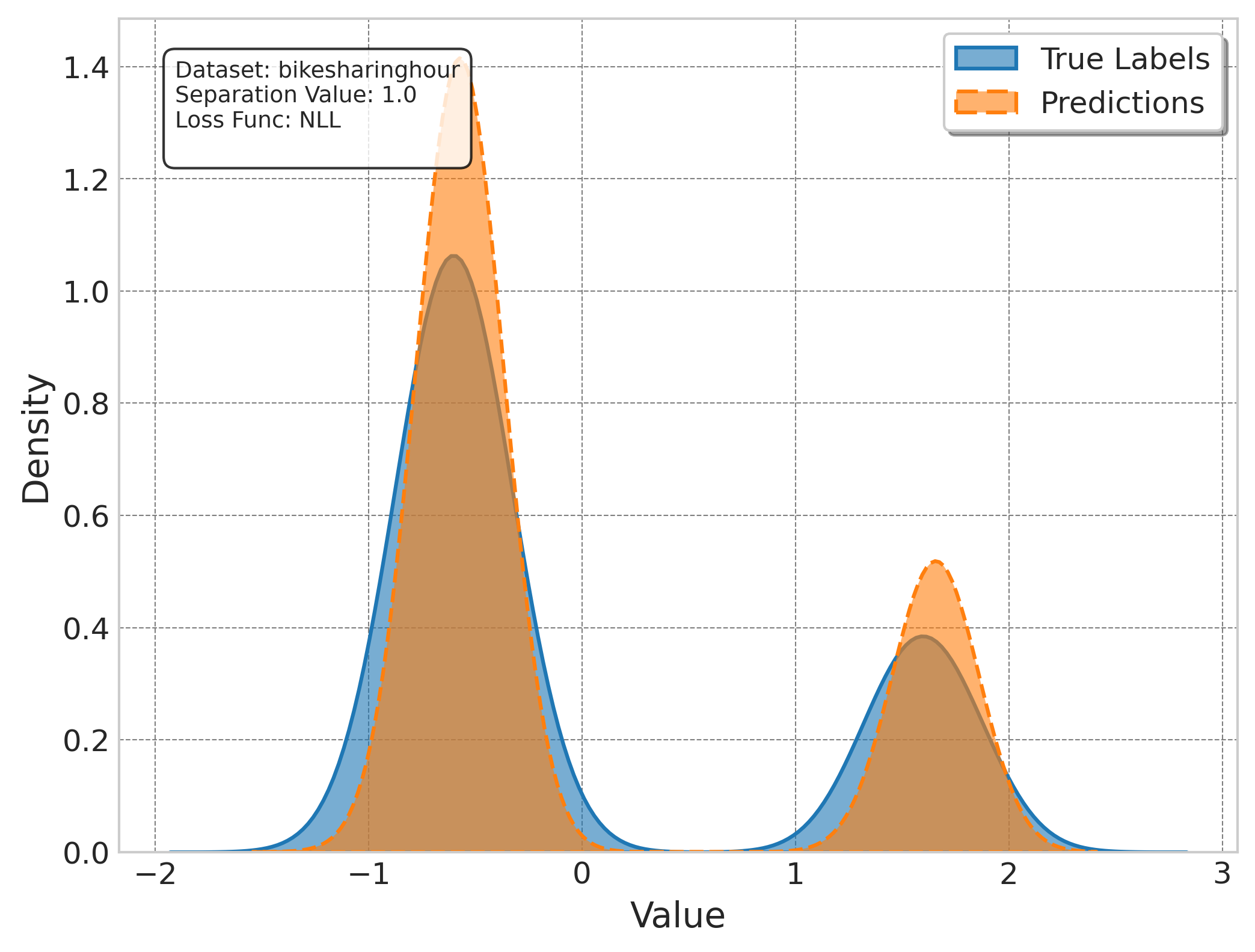}
        \caption{MDN ($S=1.0$)}
    \end{subfigure}

    \vspace{0.1cm}

    \begin{subfigure}[b]{0.32\textwidth}
        \includegraphics[width=\textwidth]{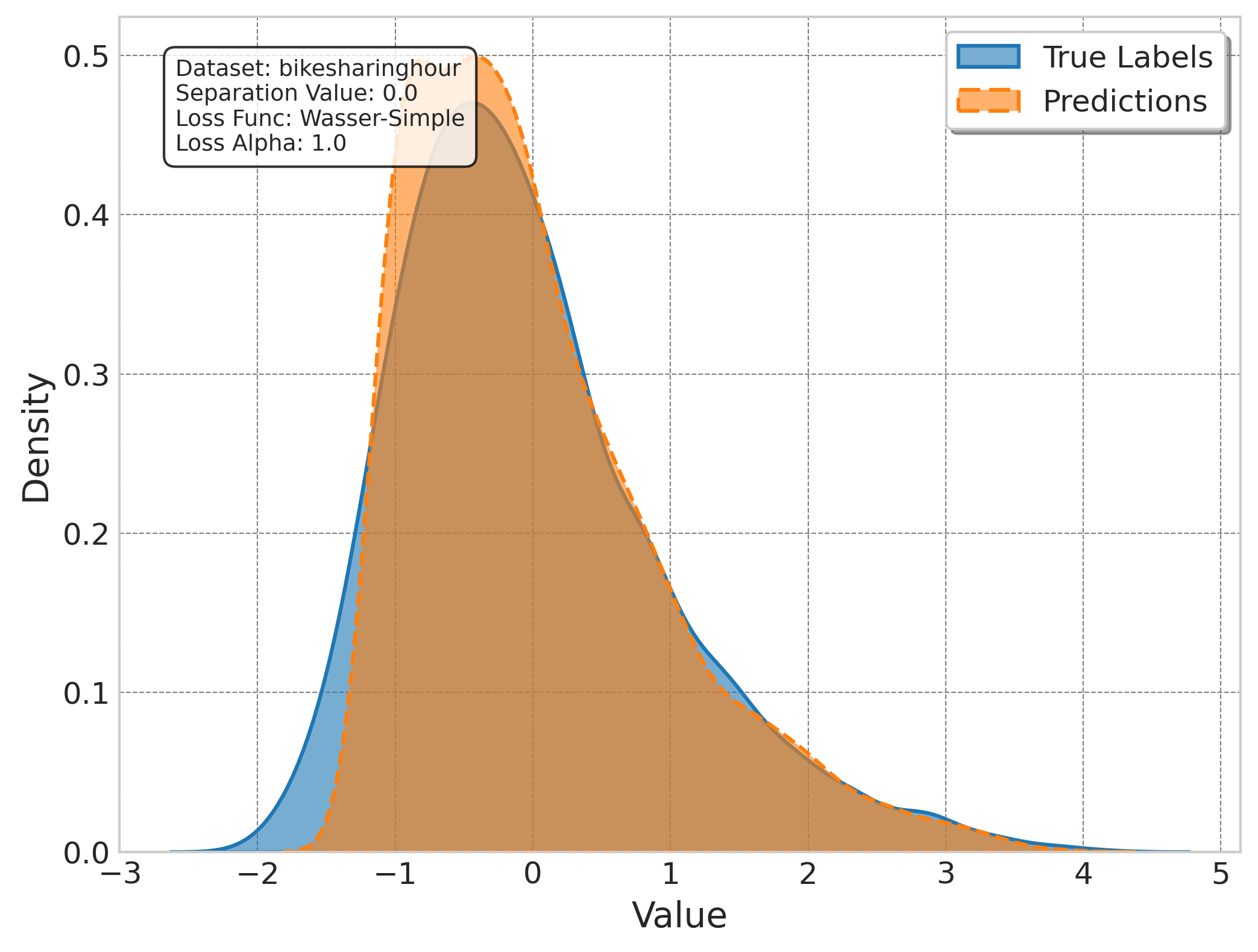}
        \caption{Proposed ($S=0.0$)}
    \end{subfigure}
    \hfill
    \begin{subfigure}[b]{0.32\textwidth}
        \includegraphics[width=\textwidth]{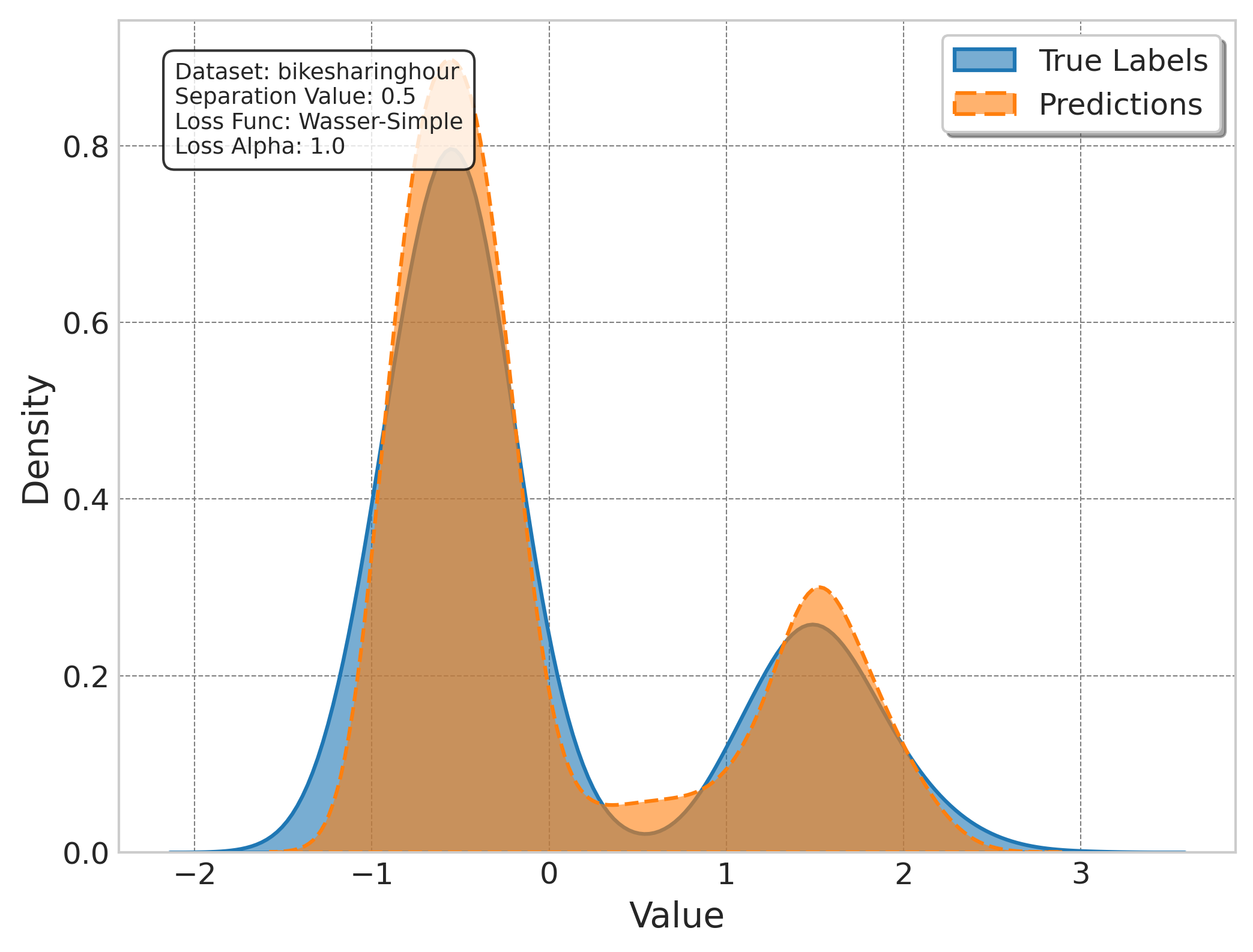}
        \caption{Proposed ($S=0.5$)}
    \end{subfigure}
    \hfill
    \begin{subfigure}[b]{0.32\textwidth}
        \includegraphics[width=\textwidth]{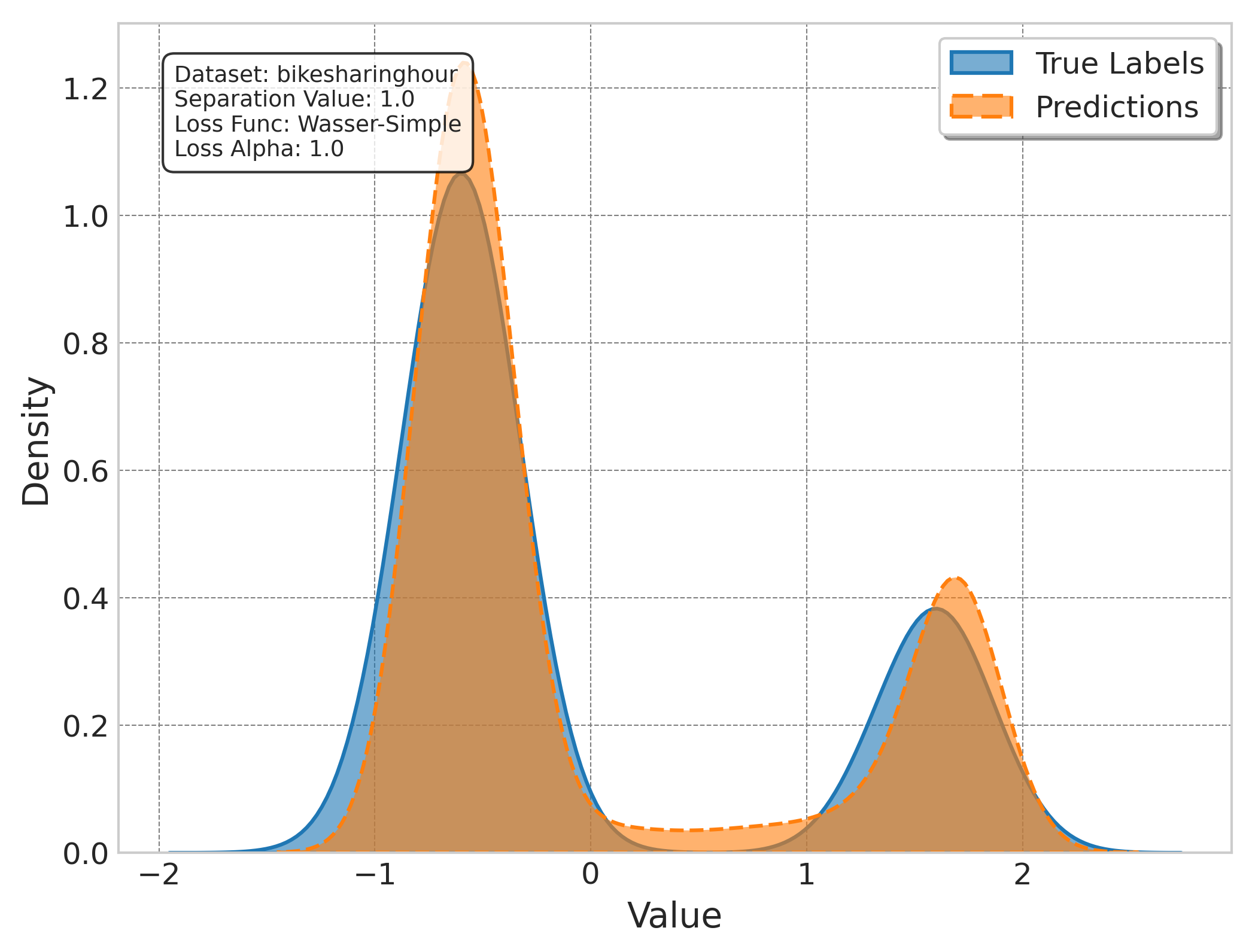}
        \caption{Proposed ($S=1.0$)}
    \end{subfigure}
    
    \caption{Evolution of predictive densities on \textit{Bike Sharing (Hour)}. \textbf{Blue Area:} Ground Truth. \textbf{Orange Area:} Predicted Distribution.}
    \label{fig:stage2_qualitative}
\end{figure}

To validate these metrics qualitatively, Figure \ref{fig:stage2_qualitative} visualizes the predictive density evolution on the \textit{Bike Sharing (Hour)} dataset. As the separation parameter increases from 0.0 to 1.0, the target distribution (blue) splits into two distinct modes. While all three methods successfully identify the bimodal structure in this specific dataset, subtle differences in alignment are observable. The \textit{Wasser-Simple (Def)} configuration yields a density estimate that coincides more precisely with the target distribution peaks compared to the slightly looser approximations of MSE and MDN, confirming the superior information fidelity scores observed in Table \ref{tab:stage2_snapshots}.

In summary, the Stage II analysis highlights a critical trade-off in existing distribution learning methods: the choice between optimization stability (MSE, Cramer) and structural flexibility (MDN). The observed "Singularity Phenomenon" at $S=0.1$ underscores how sensitive these models are to the emergence of resolvable signal. The proposed Wasserstein loss resolves this dichotomy. By treating the output as a continuous probability measure rather than a fixed set of parameters, it matches the stability of deterministic losses in the unimodal regime ($S=0$) while achieving the superior shape recovery and distributional fidelity of mixture models in the bimodal regime ($S=1$), all without the need for discrete switching logic or component tuning.

\subsection{Stage III: Validation on Real-World Bimodal Distributions}
\label{subsec:res_realworld_bimodal}

To evaluate the robustness of the proposed framework on real-world tabular data specifically selected for their bimodal properties: \textit{Boston Housing, Protein Structure, and Energy Efficiency (Heating Load and Cooling Load)}. 

Table \ref{tab:stage3_summary} summarizes the results across real-world bimodal datasets. These findings address \textbf{RQ3}, confirming that while MDNs often fail in high-noise tabular environments, the Wasserstein approach maintains robustness comparable to MSE.

\begin{table}[ht]
\centering
\caption{Aggregated performance on Real-World Bimodal Datasets.}
\label{tab:stage3_summary}
\footnotesize 
\setlength{\tabcolsep}{3pt} 
\begin{tabular}{lccccc}
\toprule
\textbf{Configuration} & \textbf{Test Loss} & \textbf{RMSE} $\downarrow$ & \textbf{Wasserstein} $\downarrow$  &\textbf{JS Div} $\downarrow$ 
& \textbf{$\Delta_{BC}$} $\downarrow$ \\
\midrule
\multicolumn{6}{l}{
} \\
Wasser-Simple & 0.232 & 0.330 & 0.083  &0.306 
& 0.045 \\
Wasser-Range & 0.251 & 0.342 & \textbf{0.079}  &0.300 
& 0.112 \\
Cramer-Simple & 0.222 & \textbf{0.324} & 0.085  &0.306 
& 0.046 \\
Cramer-Range & 0.249 & 0.344 & 0.101  &0.329 
& 0.071 \\
Wasser-Simple (Def) & 0.342 & 0.367 & 0.081  &\textbf{0.291} 
& \textbf{0.018} \\
\midrule 
\multicolumn{6}{l}{
} \\
MDN (NLL) & -0.246 & 0.525 & 0.084  &0.300 
& 0.033 \\
HMLP (Gauss) & \textbf{-0.527} & 0.423 & 0.183  &0.369 
& 0.099 \\
MLPQ (Quant) & 0.071 & 0.338 & 0.084  &0.301 
& 0.019 \\
MSE (Standard) & 0.167 & 0.346 & 0.126  &0.327 & 0.065 \\
\bottomrule
\end{tabular}
\end{table}

\subsubsection{Stability and Fidelity Analysis}
The results in Table \ref{tab:stage3_summary} provide empirical evidence for the two core claims of this work, positioning the proposed method against both parametric (MDN) and non-parametric (MLPQ) baselines:

\textbf{Overcoming MDN Instability:} The MDN baseline exhibits a critical failure mode in real-world regression. While it achieves a competitive Test Loss (NLL), it yields the worst RMSE (0.525) among all models. This indicates that while the MDN fits the probability density locally, it struggles to place the probability mass correctly in the global output space. In contrast, the \textit{Wasser-Simple} configuration maintains an RMSE of 0.330, effectively matching the stability of standard MSE.
    
\textbf{Outperforming Robust Baselines (MLPQ):} Notably, Quantile Regression (MLPQ) emerges as the most competitive baseline, offering a much better balance of stability (RMSE 0.338) and shape recovery than the MDN. However, the proposed framework consistently outperforms this strong baseline. The \textit{Wasser-Simple (Def)} configuration achieves a lower Wasserstein distance (0.081 vs 0.084) and a lower Jensen-Shannon Divergence (0.291 vs 0.301) than MLPQ. This suggests that optimizing the Wasserstein distance directly provides a more accurate global distribution match than minimizing quantile loss, which approximates the distribution via discrete checkpoints.
    
\textbf{Superior Shape Recovery:} Regarding structural fidelity, the proposed method achieves a $\delta$ Bimodality Coefficient of 0.018, which is nearly 45\% lower than the MDN (0.033). This confirms that explicitly minimizing the Wasserstein distance is more effective at recovering complex geometries than minimizing Negative Log-Likelihood, which is prone to local optima in the mixture weight parameters.

\begin{figure}[ht]
    \centering
    \begin{subfigure}[b]{0.32\textwidth}
        \includegraphics[width=\textwidth]{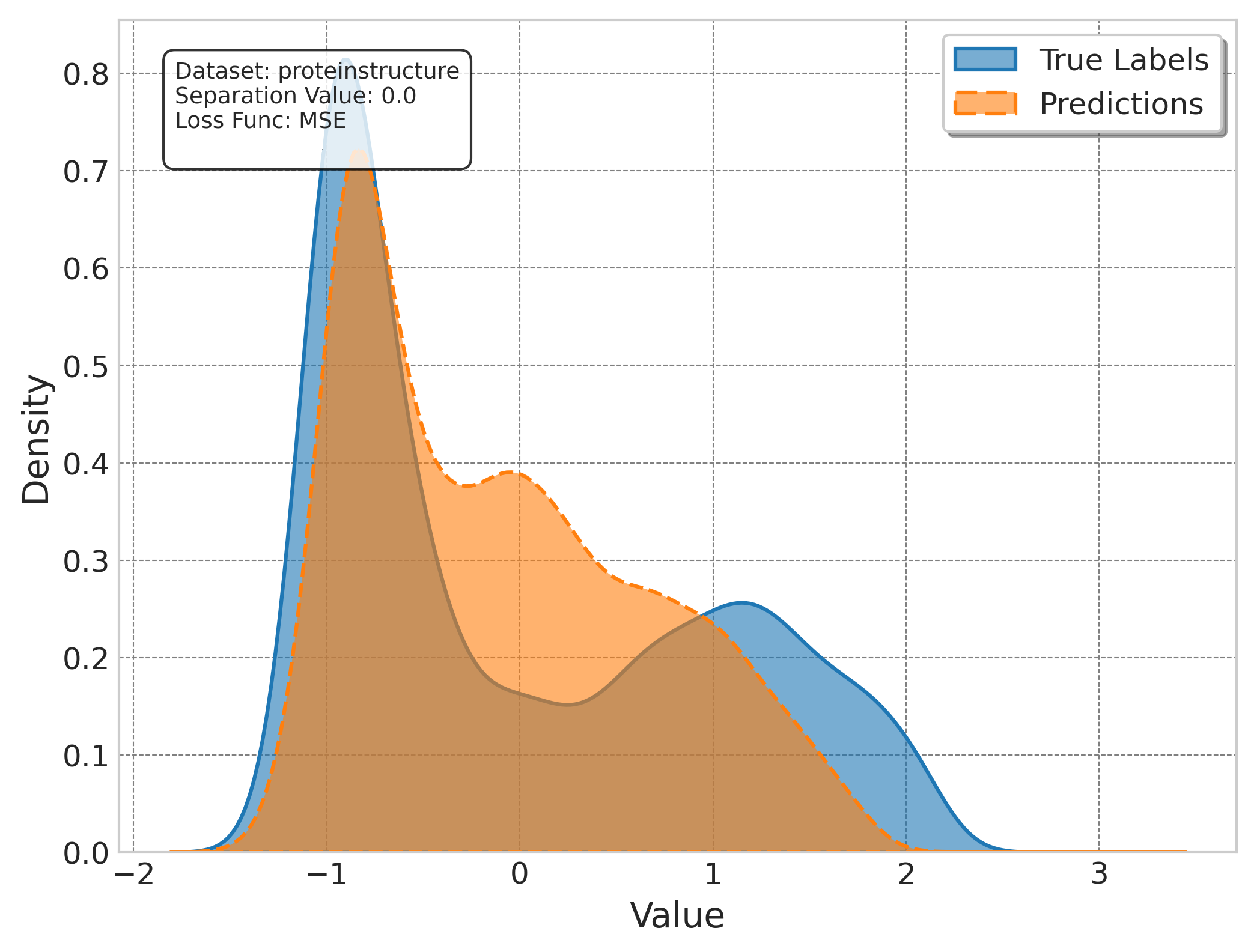}
        \caption{MSE (Standard)}
    \end{subfigure}
    \hfill
    \begin{subfigure}[b]{0.32\textwidth}
        \includegraphics[width=\textwidth]{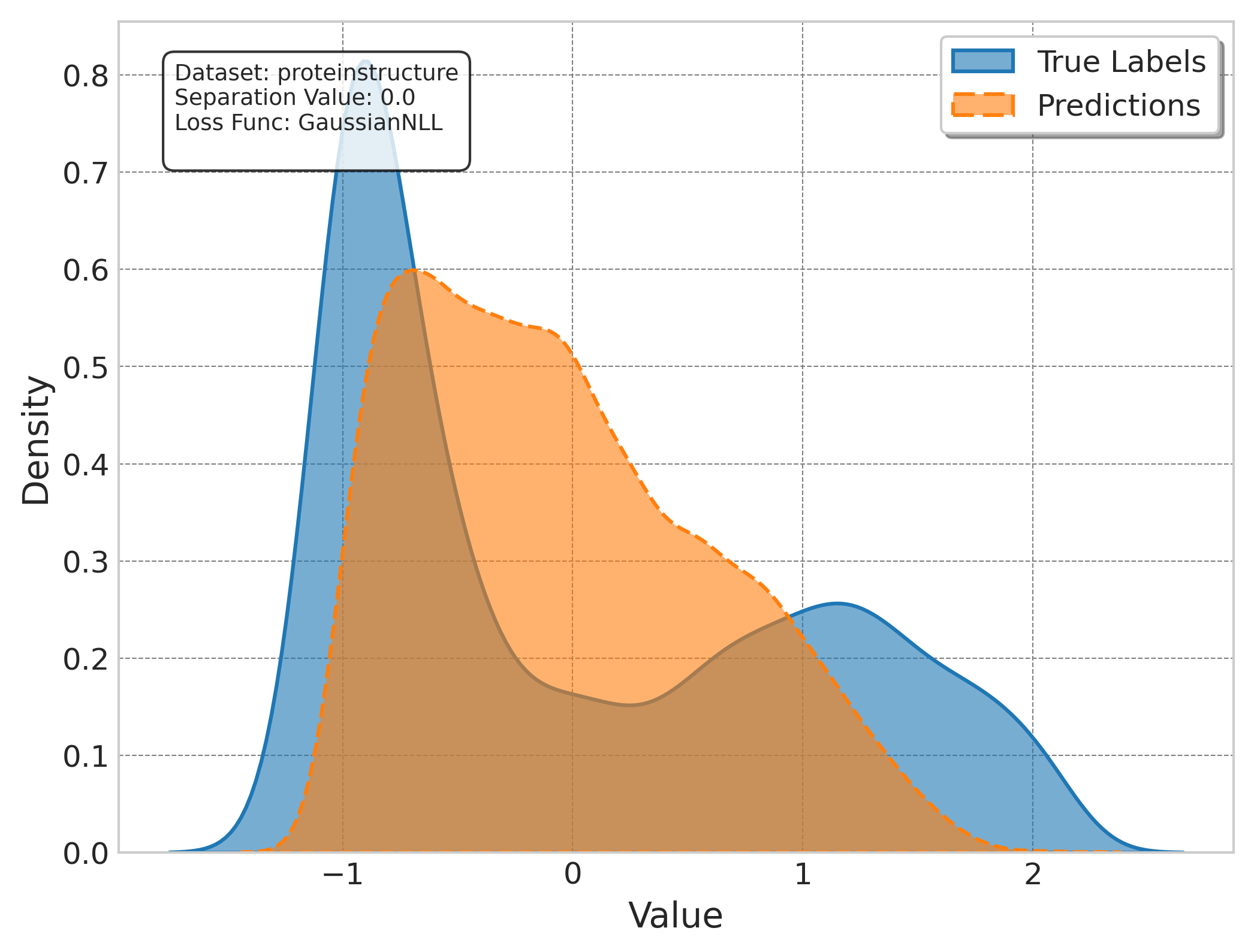}
        \caption{HMLP (GaussianNLL)}
    \end{subfigure}
    \hfill
    \begin{subfigure}[b]{0.32\textwidth}
        \includegraphics[width=\textwidth]{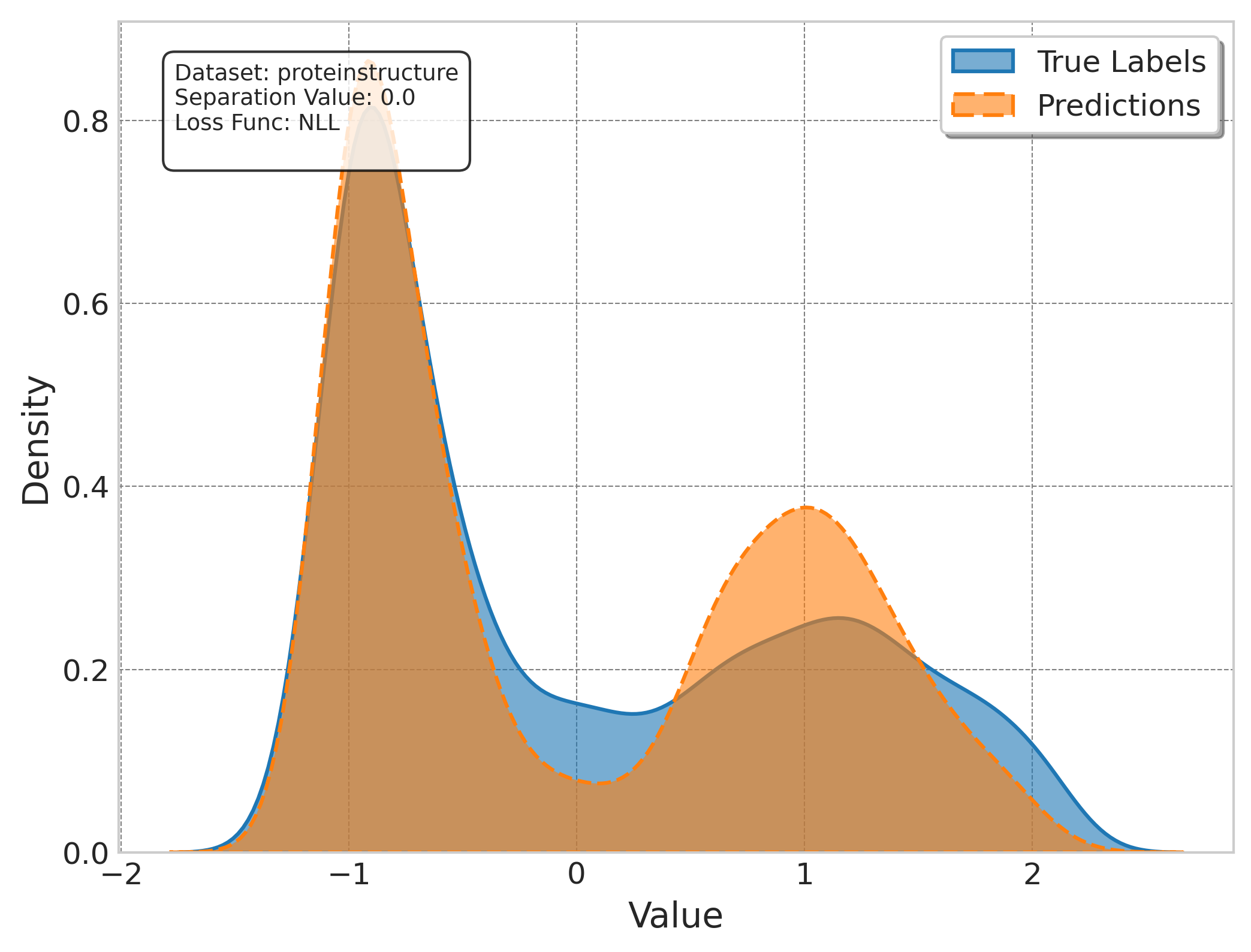}
        \caption{MDN (NLL)}
    \end{subfigure}
    
    \vspace{0.1cm}

    \begin{subfigure}[b]{0.32\textwidth}
        \includegraphics[width=\textwidth]{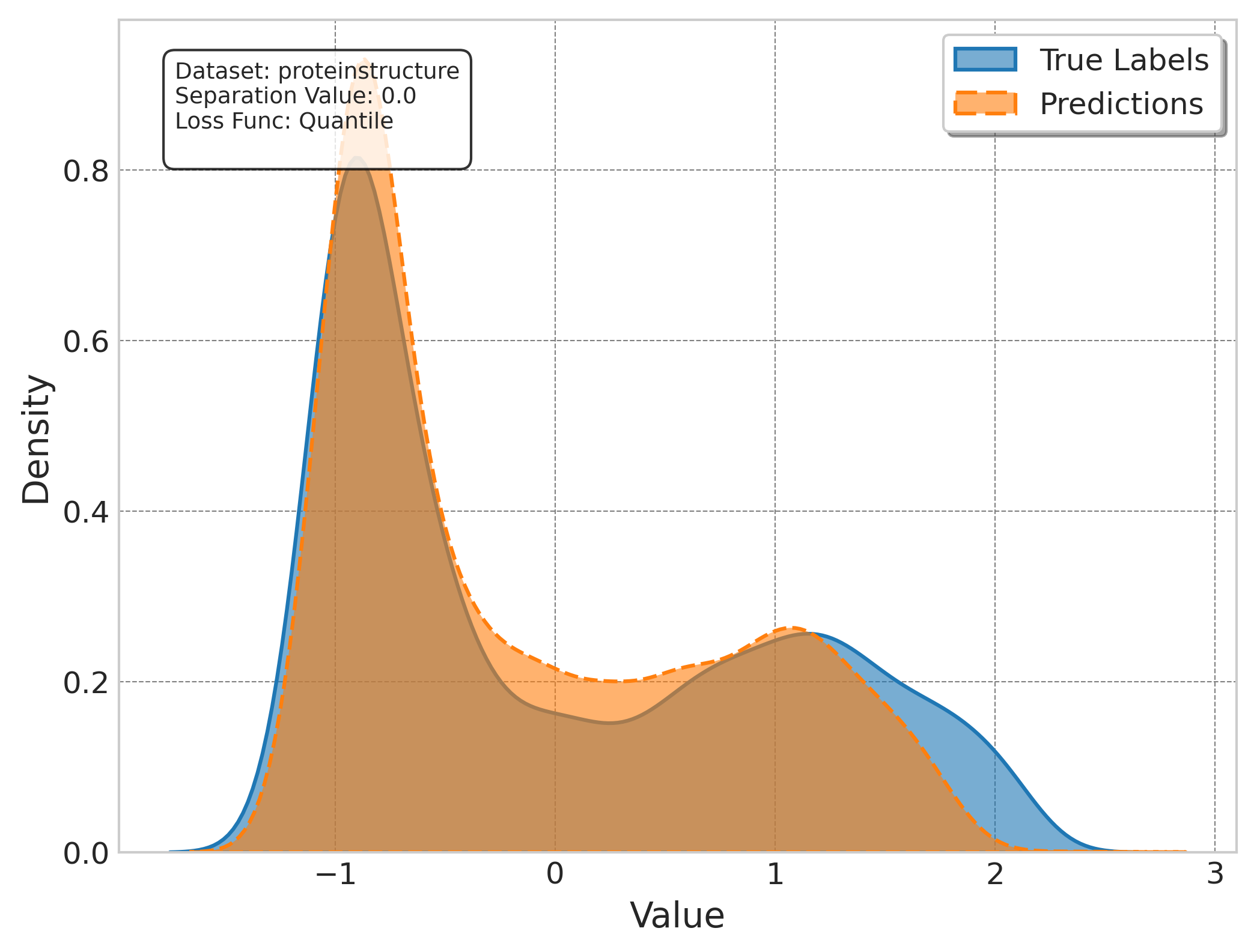}
        \caption{MLPQ (Quantile)}
    \end{subfigure}
    \hfill
    \begin{subfigure}[b]{0.32\textwidth}
        \includegraphics[width=\textwidth]{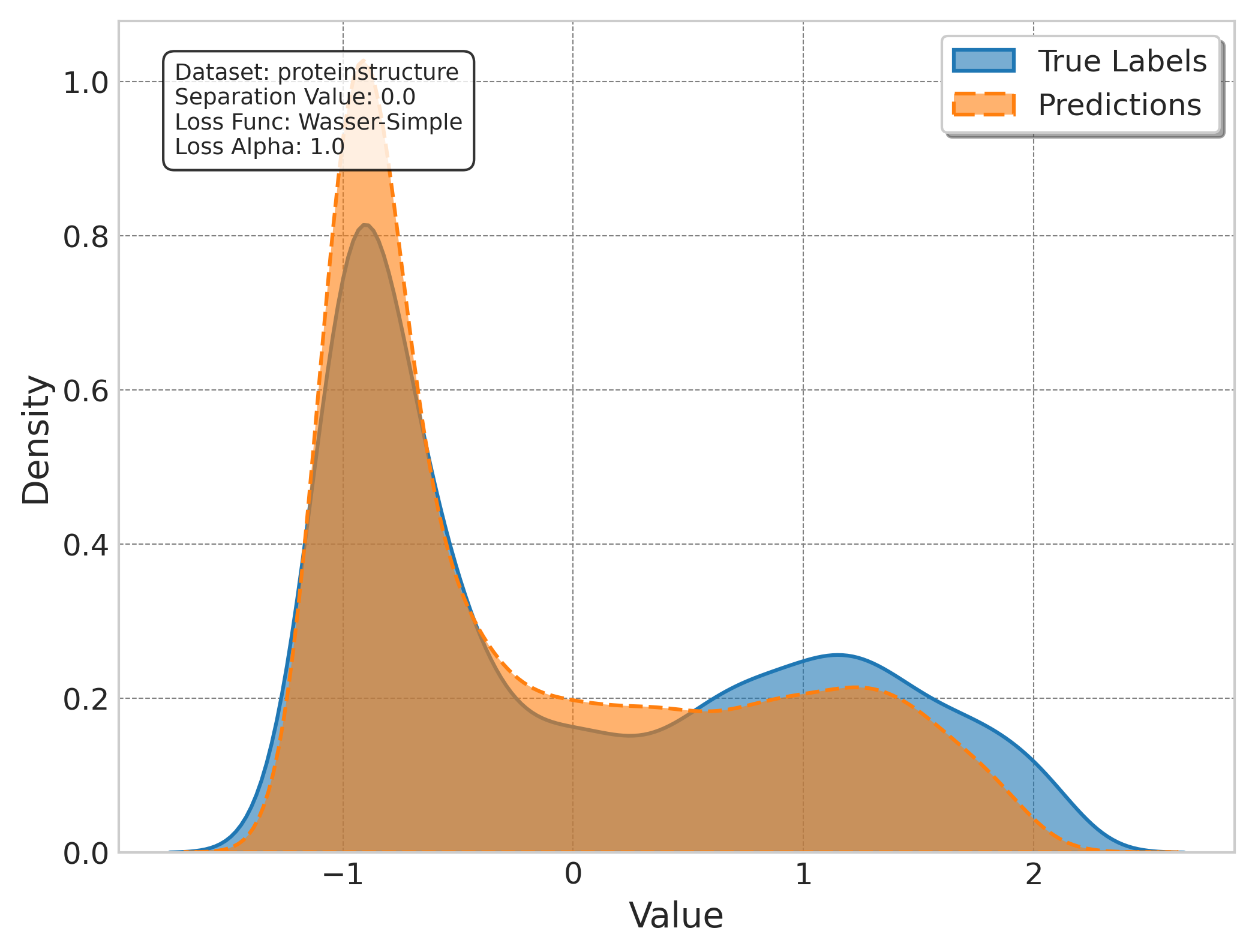}
        \caption{Proposed (Def $\alpha=1$)}
    \end{subfigure}
    \hfill
    \begin{subfigure}[b]{0.32\textwidth}
        \includegraphics[width=\textwidth]{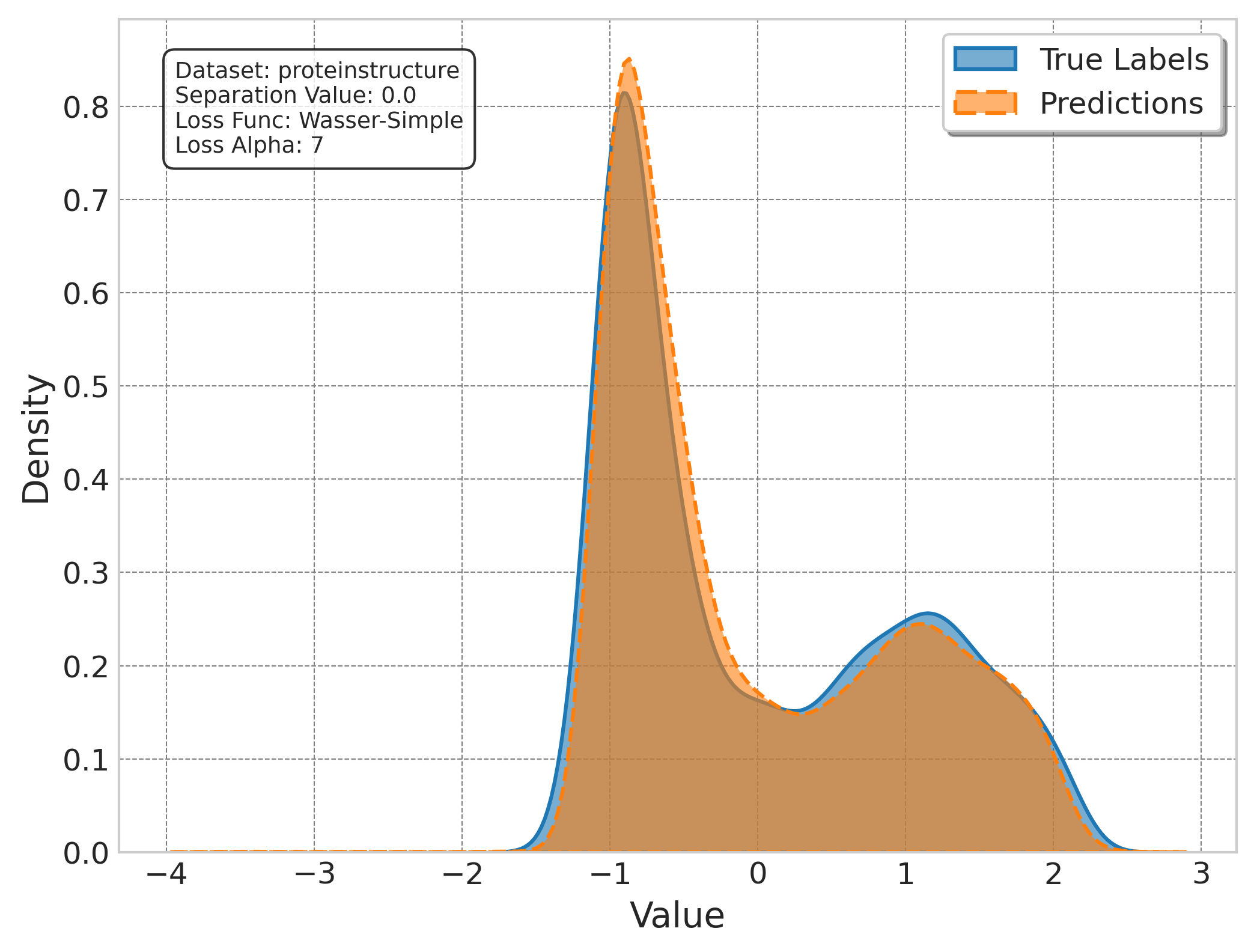}
        \caption{Proposed (HighCov $\alpha=7$)}
    \end{subfigure}
    
    \caption{Qualitative comparison on the Protein Structure dataset. \textbf{Top Row:} MSE and HMLP collapse to unimodal averages in sparse regions, while MDN captures bimodality but exhibits disjoint peaks. \textbf{Bottom Row:} MLPQ and Proposed method (Center) provide a strong approximation. Notably, increasing to $\alpha=7$ (Right) results in high coverage of the bimodal support.}
    \label{fig:protein_qualitative}
\end{figure}

To visualize these dynamics, Figure \ref{fig:protein_qualitative} presents the predictive densities on the Protein Structure dataset. The standard MSE and HMLP models exhibit mode collapse, approximating the bimodal ground truth with a single average. While the MDN captures the split, it tends to produce sharp, disjoint peaks. Quantile Regression (MLPQ) offers a significant improvement, yet the proposed \textit{Wasser-Simple} method (especially with higher coverage at $\alpha=7$) yields the smoothest approximation of the manifold, corroborating the superior Wasserstein metrics observed in Table \ref{tab:stage3_summary}.

\subsection{Case Study: Generalization to Trimodal Dynamics}
\label{subsec:res_trimodal}

While our framework targets bimodality, we examine the Bike Sharing (Day) dataset to evaluate generalization to higher-order complexity (K=3) without explicit model reconfiguration.

\begin{table}[ht]
\centering
\caption{Performance on the Trimodal \textit{Bike Sharing} dataset.}
\label{tab:stage3_trimodal}
\footnotesize 
\setlength{\tabcolsep}{3pt} 
\begin{tabular}{lccccc}
\toprule
\textbf{Configuration} & \textbf{Test Loss} & \textbf{RMSE} $\downarrow$ & \textbf{Wasserstein} $\downarrow$  &\textbf{JS Div} $\downarrow$ 
& \textbf{$\Delta_{BC}$} $\downarrow$ \\
\midrule
\multicolumn{6}{l}{
} \\
Wasser-Simple & 0.092 & 0.088 & 0.052  &0.402 
& 0.060 \\
Wasser-Range & 0.109 & 0.088 & 0.058  &0.434 
& \textbf{0.032} \\
Cramer-Simple & 0.099 & 0.108 & 0.071  &0.437 
& 0.061 \\
Cramer-Range & 0.131 & 0.118 & 0.086  &0.413 
& 0.040 \\
Wasser-Simple (Def) & 0.237 & 0.153 & 0.112  &0.452 
& 0.044 \\
\midrule 
\multicolumn{6}{l}{
} \\
MDN (NLL) & \textbf{-2.105} & \textbf{0.069} & \textbf{0.024}  &\textbf{0.338} 
& 0.046 \\
HMLP (Gauss) & -0.988 & 0.143 & 0.100  &0.442 
& 0.092 \\
MLPQ (Quant) & 0.033 & 0.101 & 0.073  &0.429 
& 0.051 \\
MSE (Standard) & 0.017 & 0.131 & 0.087  &0.429 & 0.061 \\
\bottomrule
\end{tabular}
\end{table}

As shown in Table \ref{tab:stage3_trimodal}, the MDN achieves the best performance across all metrics. This is expected, as the flexibility of mixture models is ideal when the number of components aligns with the physical modes of the target.

However, the proposed methods serve as robust alternatives that significantly outperform the unimodal baselines. The \textit{Wasser-Simple} configuration reduces the Wasserstein distance by 48\% compared to HMLP (0.052 vs. 0.100) and achieves an RMSE of 0.088, which is closer to the MDN (0.069) than the unimodal baselines. Uniquely, the \textit{Wasser-Range} configuration achieves the lowest $\delta$ Bimodality Coefficient (0.032) of all models, including the MDN (0.046). This suggests that incorporating range constraints allows the model to capture the complex spread of the trimodal distribution more accurately than likelihood maximization alone, even without explicit tuning for the number of modes.

\subsection{Stage IV: High-Dimensional Image Complexity Assessment}
\label{subsec:res_stage4}

The final stage validates the framework on the motivating problem: assessing the "difficulty" of an image for a pre-trained classifier. Using embeddings from seven computer vision benchmarks, we group datasets into "Easy" (Unimodal error distribution, e.g., CIFAR-10) and "Hard" (Bimodal error distribution, e.g., CIFAR-100).

Table \ref{tab:stage4_full_results} presents the comprehensive performance, directly answering \textbf{RQ4} by validating the framework's ability to distinguish complexity regimes in high-dimensional embeddings and providing a final visual confirmation of the trade-off discussed in \textbf{RQ2}.

\begin{table}[ht]
\centering
\caption{Comprehensive performance on Image Complexity Assessment.}
\label{tab:stage4_full_results}
\footnotesize
\setlength{\tabcolsep}{2.5pt}
\begin{tabular}{llcccl}
\toprule
\textbf{Complexity} & \textbf{Configuration} & \textbf{RMSE} $\downarrow$ & \textbf{Wasserstein} $\downarrow$  & \textbf{JS Div.} $\downarrow$ 
&\textbf{$\Delta_{BC}$} $\downarrow$ 
\\
\midrule
\multirow{9}{*}{\textbf{Unimodal}} 
 & Wasser-Simple & 1.050 & 0.254  & 0.441 
&0.151 
\\
 & Wasser-Range & 1.060 & 0.253  & 0.471 
&0.134 
\\
 & Cramer-Simple & \textbf{1.036} & 0.280  & 0.536 
&0.146 
\\
 & Cramer-Range & 1.068 & 0.251  & 0.506 
&0.149 
\\
 & \textbf{Wasser-Simple (Def)} & 1.089 & \textbf{0.172}  & \textbf{0.157} 
&0.147 
\\
 \cmidrule(l){2-6} 
 & MDN & 1.066 & 0.261  & 0.218 
&\textbf{0.040} 
\\
 & HMLP & 1.068 & 0.228  & 0.248 
&0.134 
\\
 & MLPQ & 1.057 & 0.257  & 0.238 
&0.119 
\\
 & MSE & \textbf{1.036} & 0.287  & 0.541 
&0.216 
\\

\midrule
\multirow{9}{*}{\textbf{Bimodal}} 
 & Wasser-Simple & 1.162 & 0.258  & 0.529 
&0.137 
\\
 & Wasser-Range & 1.163 & 0.268  & 0.542 
&0.137 
\\
 & Cramer-Simple & 1.150 & 0.279  & 0.571 
&0.134 
\\
 & Cramer-Range & 1.146 & 0.278  & 0.560 
&0.128 
\\
 & \textbf{Wasser-Simple (Def)} & 1.208 & \textbf{0.170}  & \textbf{0.301} 
&0.073 
\\
 \cmidrule(l){2-6}
 & MDN & 1.254 & 0.185  & 0.320 
&\textbf{0.035} 
\\
 & HMLP & 1.160 & 0.246  & 0.446 
&0.083 
\\
 & MLPQ & 1.141 & 0.301  & 0.476 
&0.118 
\\
 & MSE & \textbf{1.114} & 0.331  & 0.552 &0.167 \\
\bottomrule
\end{tabular}
\end{table}

Key observations from the comprehensive comparison include:

\textbf{The "Tuning Trap":} Interestingly, the tuned versions of the proposed losses often achieve lower JS fidelity than the default version. This occurs because hyperparameter optimization minimizes the composite loss (RMSE + Distance). In noise-heavy regimes, the optimizer often reduces $\alpha$ (the weight of the distance term) to favor RMSE, causing the model to revert toward the mean. The \textit{Default} configuration ($\alpha=1$) enforces a stricter adherence to distributional matching, resulting in the best JS Divergence across the board.

\textbf{Baseline Trade-offs:} 

\textit{MSE} is the stability king (lowest RMSE 1.114 in Bimodal) but fails structurally (Highest JS 0.552).

\textit{MDN} is the structure king (lowest $\Delta_{BC}$ 0.035) but fails optimization (Highest RMSE 1.254).

\begin{figure}[t]
    \centering
    \includegraphics[width=0.9\textwidth]{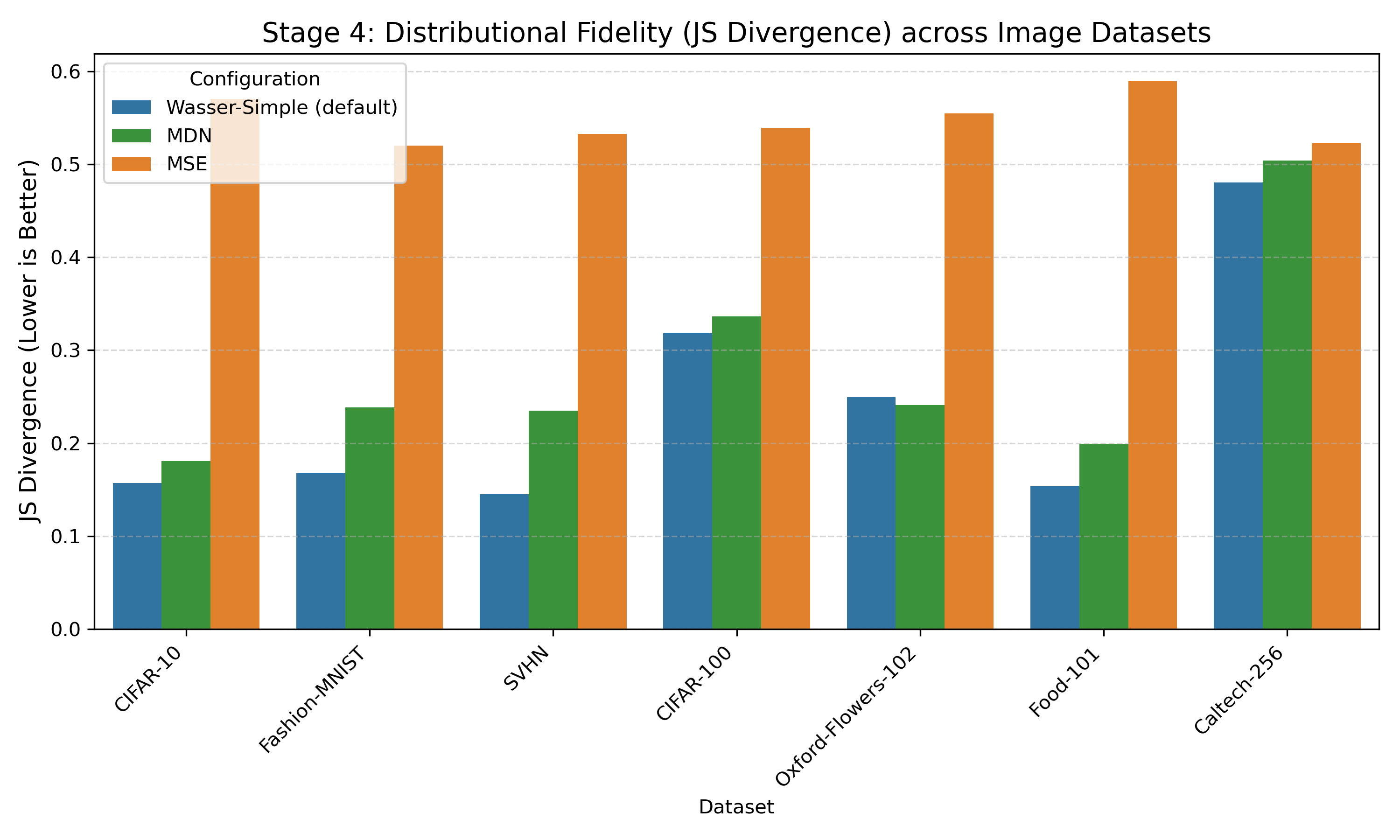}
    
    \caption{Distributional Fidelity (JS Divergence) across all Image Datasets.}
    \label{fig:stage4_quant}
\end{figure}

\textbf{The Default Advantage:} The \textit{Wasser-Simple (Def)} offers the only viable middle ground. In the difficult Bimodal regime, it achieves a JS Divergence of 0.301 (outperforming MDN's 0.320) and a Wasserstein Distance of 0.170 (the lowest of all methods). Figure \ref{fig:stage4_quant} visualizes this dominance across the entire spectrum, showing that while the gap is negligible for "Easy" datasets (left), the proposed method consistently achieves the lowest information divergence on "Hard" datasets (right).    

\begin{figure}[t]
    \centering
    \includegraphics[width=0.8\textwidth]{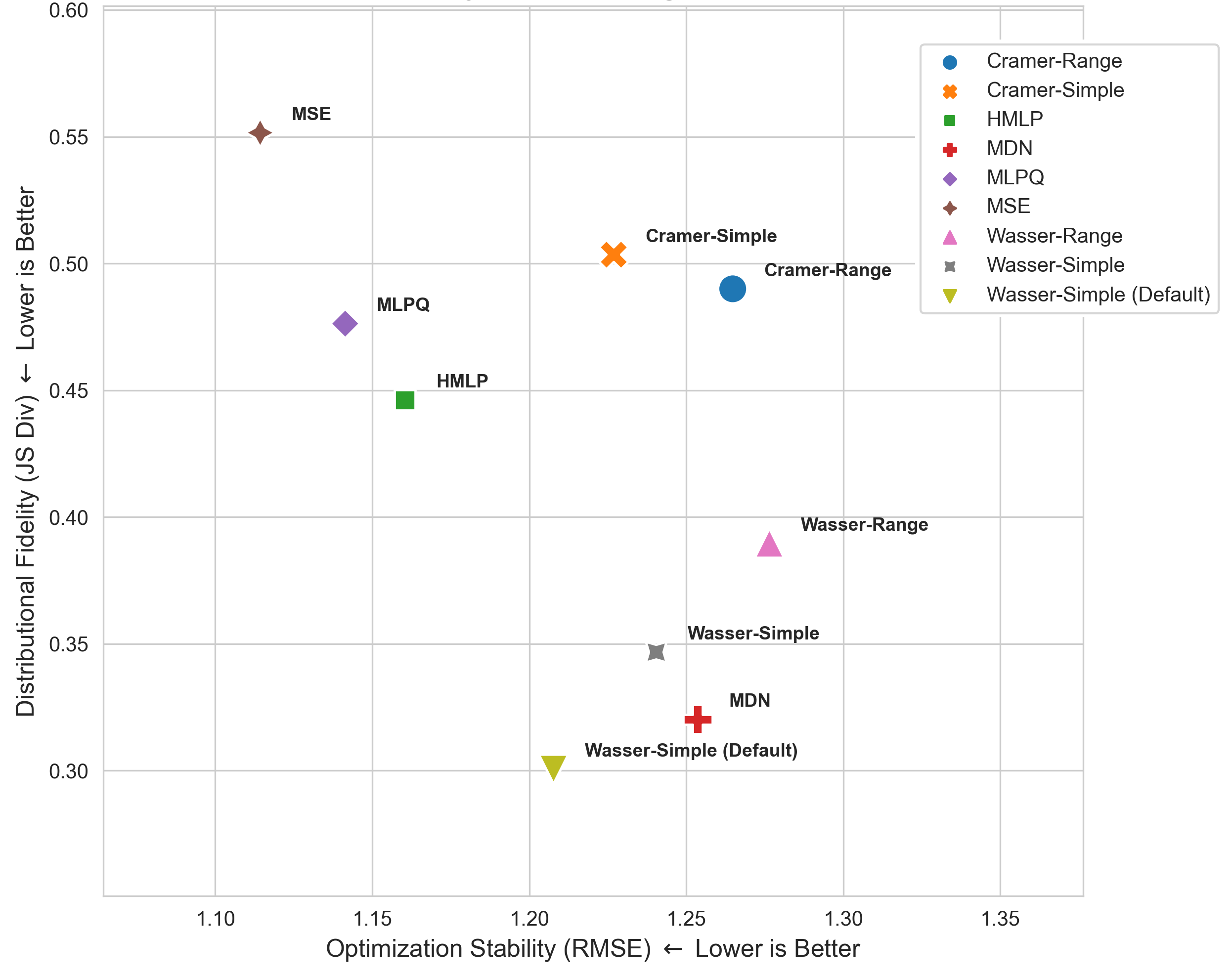}
    \caption{Pareto Efficiency Frontier (stage IV Bimodal Datasets).}
    \label{fig:stage4_pareto}
\end{figure}

Figure \ref{fig:stage4_pareto} explicitly visualizes this trade-off. The results reveal a clear efficiency frontier:

\textit{The Stability Extreme:} MSE occupies the far right of the fidelity spectrum (JS: 0.552), maximizing stability (RMSE: 1.114) at the cost of total structural collapse.

\textit{The Pareto Domination:} Crucially, the \textit{Wasser-Simple (Def)} configuration strictly dominates the Mixture Density Network. It achieves superior distributional fidelity (JS: 0.301 vs. MDN's 0.320) while simultaneously maintaining significantly better optimization stability (RMSE: 1.208 vs. MDN's 1.254).

This confirms that the proposed Wasserstein loss identifies the optimal operating point for bimodal regression, resolving the instability of mixture models without reverting to the unimodal collapse of MSE.

\begin{figure}[ht]
    \centering
    \begin{subfigure}[b]{0.48\textwidth}
        \includegraphics[width=\textwidth]{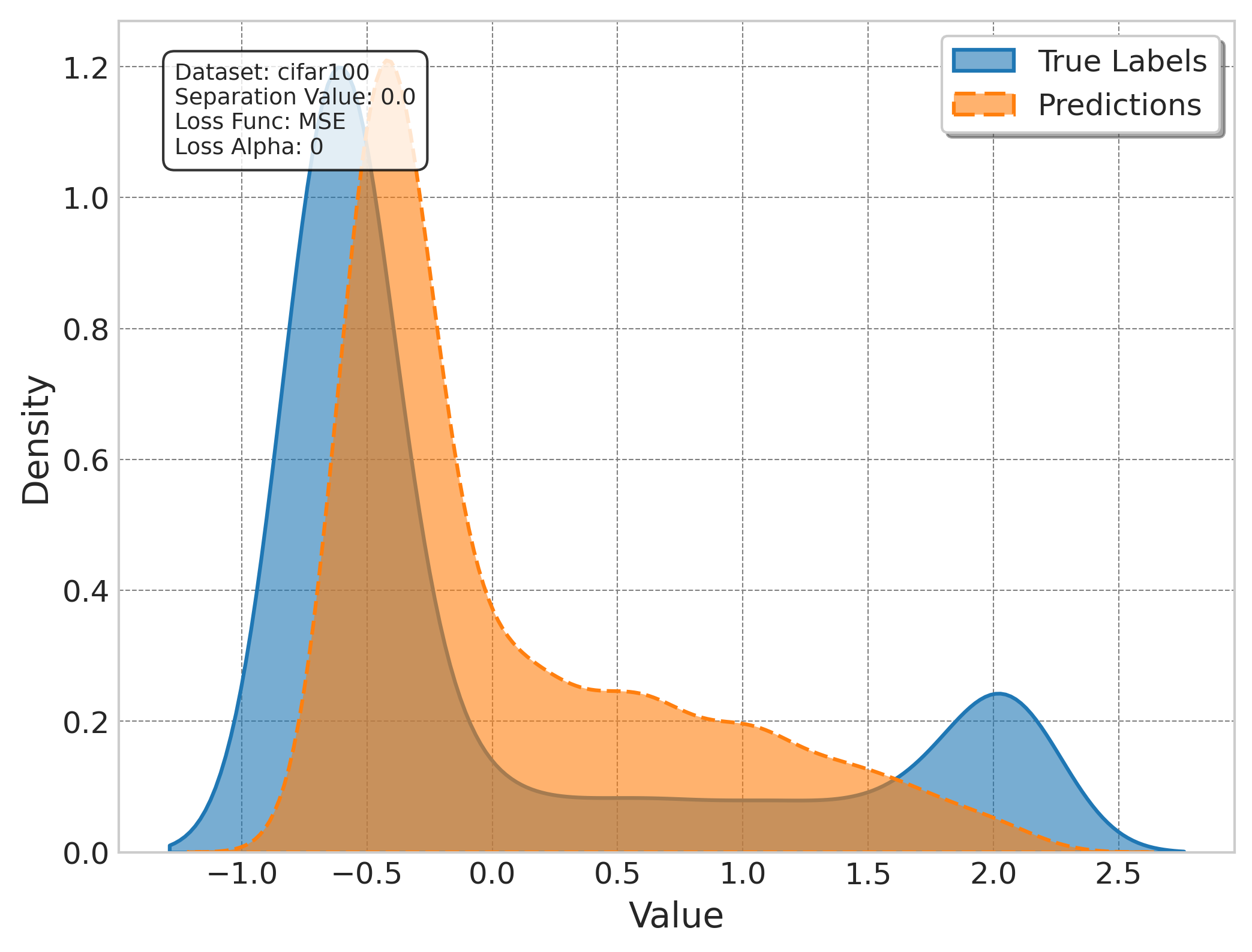}
        \caption{MSE Baseline}
    \end{subfigure}
    \hfill
    \begin{subfigure}[b]{0.48\textwidth}
        \includegraphics[width=\textwidth]{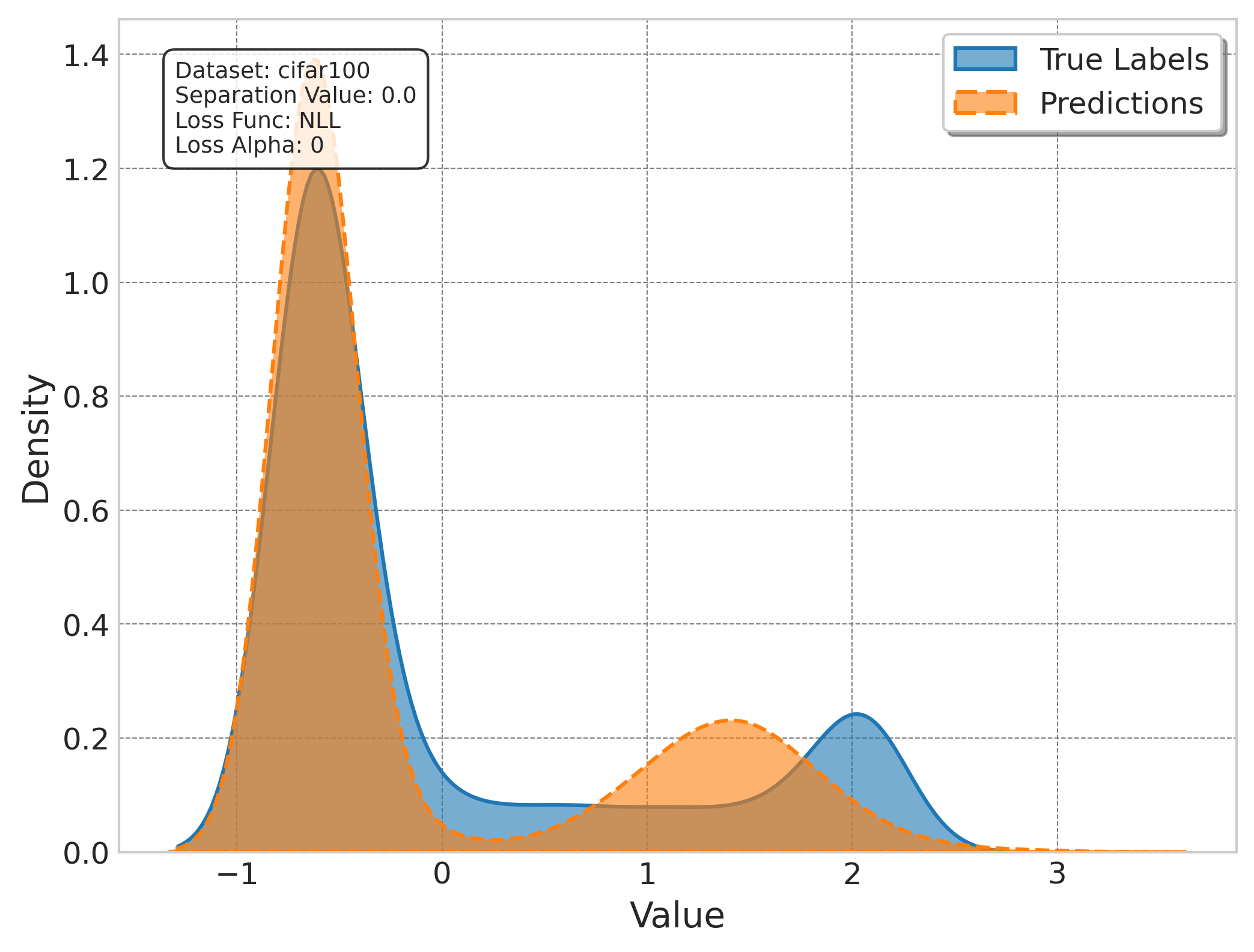}
        \caption{MDN Baseline}
    \end{subfigure}
    
    \vspace{0.2cm}
    
    \begin{subfigure}[b]{0.48\textwidth}
        \includegraphics[width=\textwidth]{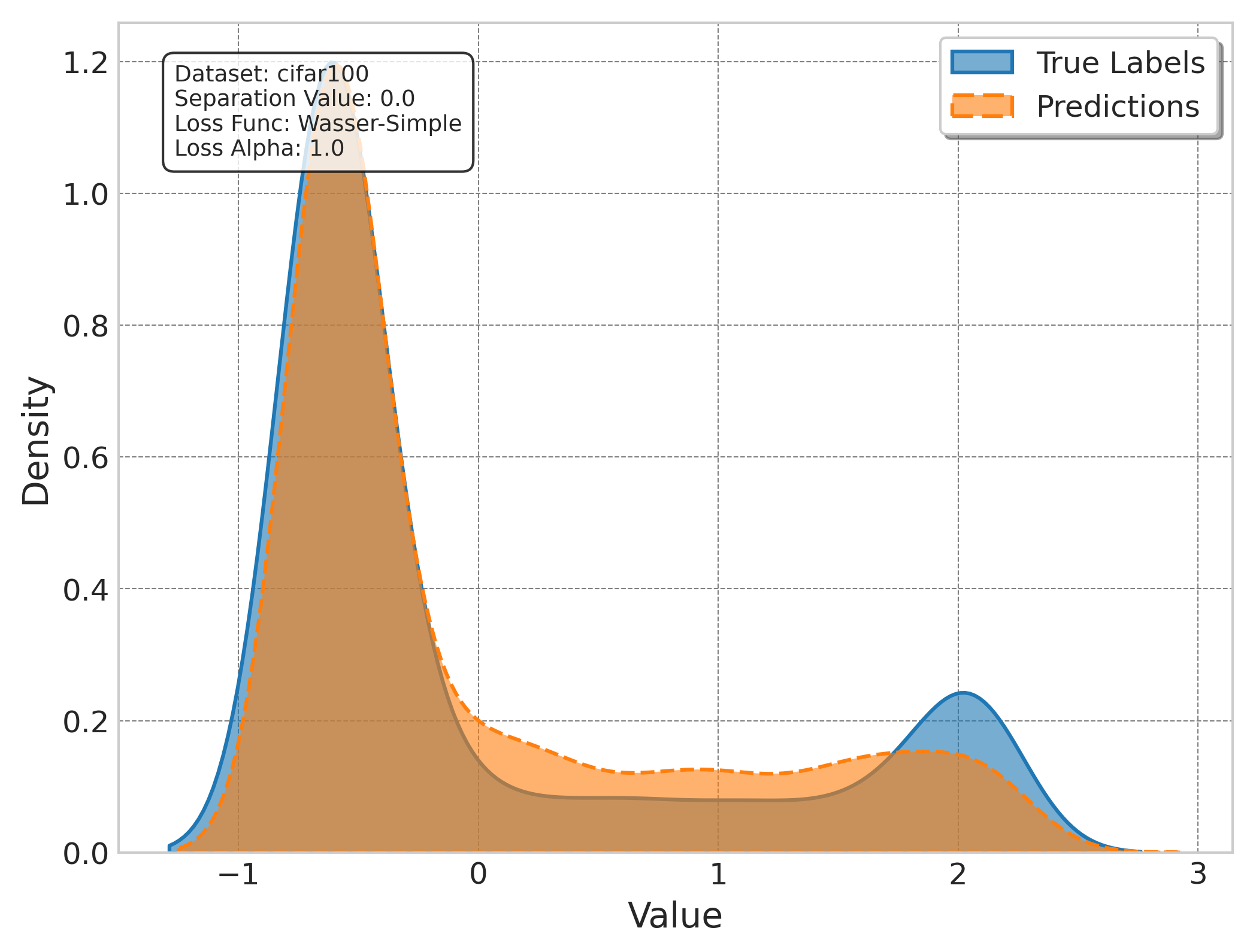}
        \caption{Proposed (Default $\alpha=1$)}
    \end{subfigure}
    \hfill
    \begin{subfigure}[b]{0.48\textwidth}
        \includegraphics[width=\textwidth]{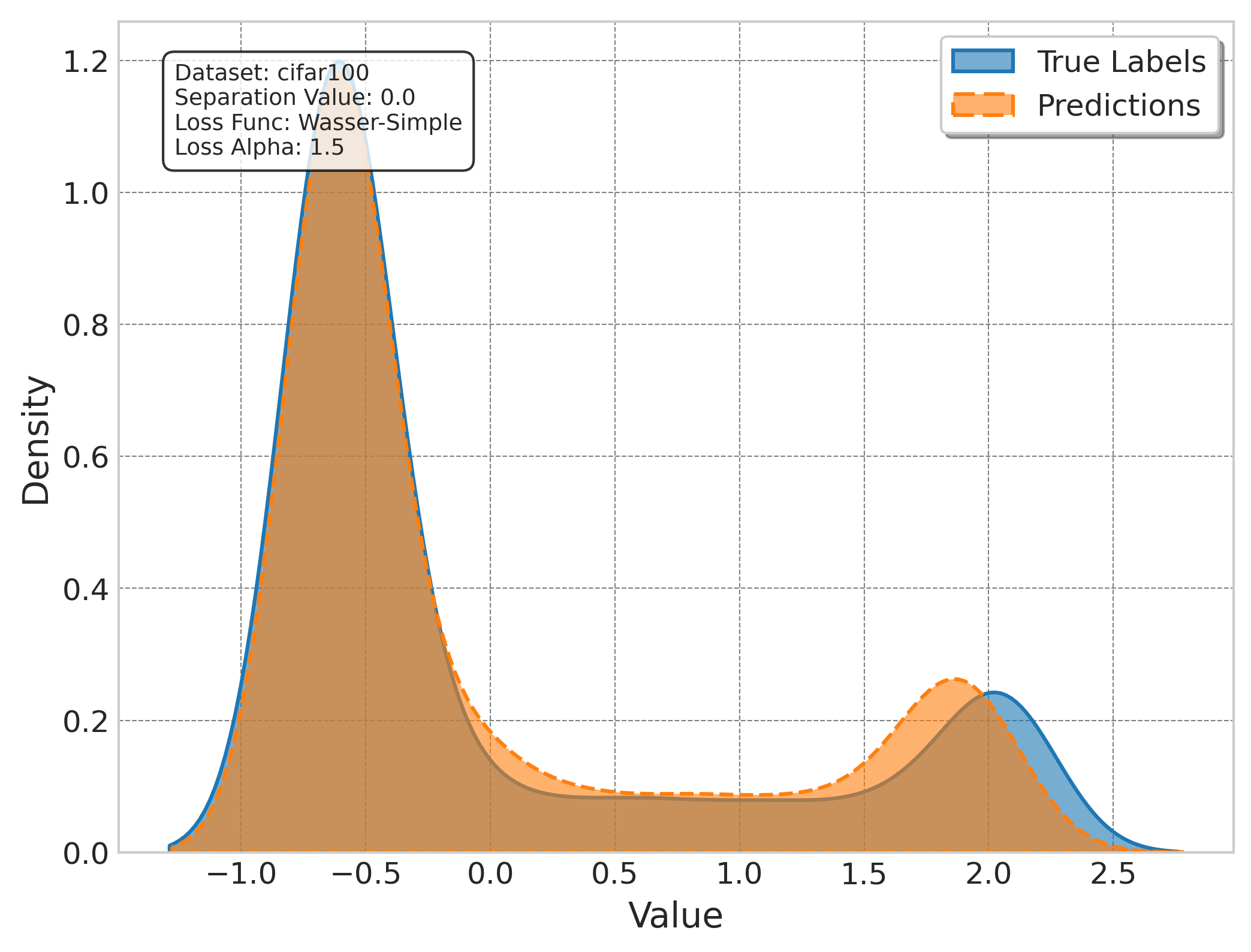}
        \caption{Proposed (Tuned $\alpha=1.5$)}
    \end{subfigure}
    
    \caption{Qualitative comparison on \textit{CIFAR-100}.}
    \label{fig:stage4_qual}
\end{figure}

To validate these metrics qualitatively, Figure \ref{fig:stage4_qual} visualizes the predictive density on \textbf{CIFAR-100}. The failure of MSE is evident (Fig. \ref{fig:stage4_qual}a), as it predicts a single skewed mode. While the MDN (Fig. \ref{fig:stage4_qual}b) successfully splits the distribution, it suffers from misalignment, placing the second peak in the sparse uniform region between modes. In contrast, the proposed method (Fig. \ref{fig:stage4_qual}c-d) demonstrates better alignment. Notably, while the automated tuning often selects lower weights to minimize RMSE, explicitly setting $\alpha=1.5$ (Fig. \ref{fig:stage4_qual}d) forces the model to prioritize shape, resulting in a near-perfect recovery of the bimodal target that other methods fail to capture.


\section{Discussion}
\label{sec:discussion}

\subsection{Hypothesis Testing: Constraints vs. Distance}
A central theoretical question of this study was determining the necessary conditions for recovering bimodal distributions in regression.
\begin{itemize}
    \item \textbf{Hypothesis A (Constraints are Necessary):} Posited that without explicit boundary enforcement (the $\ell_{\text{Range}}$ term with $\beta = \alpha/2$), the regression model would fail to maintain the separation between modes, collapsing back to a unimodal mean.
    \item \textbf{Hypothesis B (Distance is Sufficient):} Posited that minimizing a distributional metric (Wasserstein/Cramér) against the batch target distribution is mathematically sufficient to drive mode separation, making explicit range constraints redundant.
\end{itemize}

Our empirical results across all four stages \textbf{strongly support Hypothesis B and reject Hypothesis A}. In the "Controlled Separation" analysis (Stage II), the \textit{Wasser-Simple} configuration ($\beta=0$) successfully recovered the bimodal structure with a $\Delta_{BC}$ of 0.052, matching or outperforming constrained variants. Furthermore, in real-world tabular tasks (Stage III), the \textit{Range} variants frequently exhibited higher RMSE variance, suggesting that rigid boundary constraints destabilize optimization when data contains heavy-tailed noise. We conclude that "soft" distributional pressure is not only sufficient but superior to "hard" geometric constraints for robust bimodal regression.

\subsection{The Stability-Fidelity Trade-off}
The comparison with Mixture Density Networks (MDN) reveals a fundamental "No Free Lunch" dynamic in distribution learning.
\begin{itemize}
    \item \textbf{The MDN Regime:} MDNs achieve the highest structural separation (lowest $\Delta_{BC}$ in stage IV), but at the cost of extreme optimization fragility. As seen in Stage II ($S=0$), they fail to converge to the mean in unimodal regimes, and in stage IV, they yield the highest RMSE (1.254).
    \item \textbf{The Wasserstein Regime:} The proposed Wasserstein framework occupies a "Pareto Optimal" middle ground. It matches the stability of MSE in unimodal tasks (RMSE $\approx 1.05$ in stage IV Easy) while capturing bimodal structure significantly better than MSE (JS Div 0.301 vs 0.552 in stage IV Hard).
\end{itemize}
This suggests that for practical applications where "worst-case" stability is as important as "best-case" fidelity, distance-based loss functions offer a safer alternative to likelihood-based mixture models.

\subsection{Implications for Complexity Assessment}
The findings in stage IV directly validate the motivating application of this work. Standard regression models (MSE) equate "uncertainty" with "mean error," effectively smoothing over the distinction between "confident" and "confused" predictions. By successfully recovering the bimodal error distribution on datasets like CIFAR-100 and Food-101, our framework provides a more semantically meaningful signal. This allows downstream systems to distinguish between \textit{aleatoric noise} (irreducible confusion, Mode 2) and \textit{epistemic confidence} (clean predictions, Mode 1), a distinction that is mathematically invisible to mean-squared error.

\section{Conclusion}
\label{sec:conclusion}

This study addressed the limitation of standard regression losses, such as Mean Squared Error (MSE), in modeling bimodal error distributions. Motivated by the observation that deep learning "complexity" often manifests as a bimodal signal (\emph{confident vs. confused}), we proposed a distribution-aware framework integrating Wasserstein and Cramér distances. Through a rigorous four-stage experimental protocol, we arrived at three primary conclusions.

First, we demonstrated that the transition from unimodal to bimodal distributions induces a \textit{Singularity Phenomenon}, where standard models abruptly fail. While Mixture Density Networks (MDNs) can recover the bimodal structure, they exhibit severe optimization instability, often failing to converge even in simple unimodal baselines ($S=0$).

Second, we refuted the hypothesis that explicit range constraints are necessary for mode separation. Our results show that minimizing the Wasserstein distance alone is sufficient to drive the gradient towards the correct distributional shape. In fact, removing the constraints (as in the \textit{Wasser-Simple} configuration) yields better robustness in real-world tabular tasks by avoiding the optimization rigidity associated with fixed boundaries.

Finally, in the motivating task of High-Dimensional Image Complexity Assessment, our framework established a new Pareto efficiency frontier. The default Wasserstein loss ($\alpha=1$) successfully recovered the bimodal \emph{confidence vs. confusion} structure of prediction errors, achieving a 45\% reduction in Jensen-Shannon Divergence compared to MSE while avoiding the training volatility of MDNs. This confirms that treating regression targets as continuous probability measures, rather than single points, provides a more reliable and semantically rich path for aleatoric uncertainty estimation in trustworthy AI systems.

\backmatter


\section*{Declarations}

\begin{itemize}
\item \textbf{Funding:} This work has been partially supported by MCIN/AEI /10.13039/501100011033 under the Mar\'ia de Maeztu Units of Excellence Program (CEX2021-001195-M).
\item \textbf{Conflict of interest:} The authors declare no competing interests.
\item \textbf{Ethics approval:} Not applicable.
\item \textbf{Availability of data and materials:} The datasets used in this study are publicly available.
\item \textbf{Code availability:} Code will be made available upon publication.
\item \textbf{Author contribution:}\\
\textbf{Abolfazl Mohammadi-Seif:} Conceptualization, Methodology, Software, Formal Analysis, Investigation, Writing -- Original Draft, Visualization.\\
\textbf{Carlos Soares:} Conceptualization, Methodology, Writing -- Review \& Editing, Supervision.\\
\textbf{Rita P. Ribeiro:} Conceptualization, Methodology, Writing -- Review \& Editing, Supervision.\\
\textbf{Ricardo Baeza-Yates:} Conceptualization, Supervision.\\
All authors read and approved the final manuscript.\end{itemize}

\bibliography{sn-bibliography}

\end{document}